%% file: main.tex
\pdfoutput=1

\documentclass[11pt]{article}

\usepackage{acl}

\usepackage{times}
\usepackage{latexsym}

\usepackage[T1]{fontenc}

\usepackage[utf8]{inputenc}

\usepackage{microtype}

\usepackage{inconsolata}

\usepackage{bm}
\usepackage{tikz}
\usepackage{array}
\usepackage{float}
\usepackage{bbding}
\usepackage{xspace}
\usepackage{xcolor}
\usepackage{amssymb}
\usepackage{amsmath}
\usepackage{enumitem}
\usepackage{graphicx}
\usepackage{multirow}
\usepackage{multicol}
\usepackage{makecell}
\usepackage{listings}
\usepackage{setspace}
\usepackage{booktabs}
\usepackage{hyperref}
\usepackage{listings}
\usepackage{subfigure}

\usepackage[most]{tcolorbox}
\usepackage[ruled]{algorithm2e}

\setlist[itemize]{leftmargin=5mm, itemsep=0mm}

\definecolor{comment}{rgb}{0,0.6,0}
\definecolor{key}{rgb}{0.84, 0.44, 0.84}
\definecolor{func}{rgb}{0.16, 0.72, 0.86}
\definecolor{var}{rgb}{1, 0.33, 0.33}
\definecolor{true}{rgb}{1, 0.78, 0.02}

\newcommand{\ie}{\emph{i.e.,}\xspace}
\newcommand{\eg}{\emph{e.g.,}\xspace}

\newcommand{\aka}{\emph{a.k.a.,}\xspace}

\newcommand{\seq}{reflection\xspace}
\newcommand{\name}{ReflectionCoder\xspace}

\definecolor{title}{rgb}{0.2, 0.2, 0.2}
\definecolor{background}{rgb}{1, 0.975, 0.975}

\title{ReflectionCoder: Learning from Reflection Sequence for \\ Enhanced One-off Code Generation}

\author{
  Houxing Ren$^1$ \quad Mingjie Zhan$^2\footnotemark[1]$ \quad Zhongyuan Wu$^2$ \quad Aojun Zhou$^1$ \\
  \textbf{Junting Pan}$^{1,3}$ \quad \textbf{Hongsheng Li}$^{1,3,4}$\thanks{Corresponding author.} \\
  $^1$CUHK MMLab ~ $^2$SenseTime Research ~ $^3$CPII under InnoHK ~ $^4$Shanghai AI Laboratory \\
  renhouxing@gmail.com \quad zhanmingjie@sensetime.com \quad hsli@ee.cuhk.edu.hk
}

\begin{document}

\maketitle

\begin{abstract}
Code generation plays a crucial role in various tasks, such as code auto-completion and mathematical reasoning. Previous work has proposed numerous methods to enhance code generation performance, including integrating feedback from the compiler. Inspired by this, we present \name, a novel approach that effectively leverages reflection sequences constructed by integrating compiler feedback to improve one-off code generation performance. Furthermore, we propose \seq self-distillation and dynamically masked distillation to effectively utilize these reflection sequences. Extensive experiments on three benchmarks, \ie HumanEval~(+), MBPP~(+), and MultiPL-E, demonstrate that models fine-tuned with our method achieve state-of-the-art performance. Beyond the code domain, we believe this approach can benefit other domains that focus on final results and require long reasoning paths. Code and data are available at \url{https://github.com/SenseLLM/ReflectionCoder}.
\end{abstract}


\input{sec-int}
\input{sec-rel}
\input{sec-met}
\input{sec-exp}
\input{sec-con}

\bibliography{reference}

\input{sec-app}

\end{document}

%% file: sec-int.tex
\section{Introduction} \label{sec:intro}

Code generation aims to automatically produce code based on natural language description, significantly saving developers time and reducing human error. 
In the past few decades, a lot of research has been conducted for code modeling, such as CodeBert~\cite{CodeBertFengGTDFGS0LJZ20}, CodeT5~\cite{CodeT50034WJH21}.
Recently, Large Language Models~(LLMs) have shown impressive modeling ability on natural language that allows them to perform various difficult tasks~\cite{GPT42023ABS230308774}. 
By training on code domain datasets, LLMs such as CodeGen~\cite{CodeGen2023NijkampPHTWZSX23}, StarCoder~\cite{StarCoderABS230506161}, Code Llama~\cite{CodeLlamaABS230812950}, and DeepSeek-Coder~\cite{DeepSeekCoderABS240114196}, which can accurately understand user intents and generate code, have shown better performance on code-related tasks.
Leveraging this powerful capability, various works empower LLMs in complex tasks including solving mathematics problems and logic reasoning by integrating code and its execution result as Chain-of-Thoughts~(CoTs), such as PAL~\cite{PAL2023GaoMZ00YCN23} and PoT~\cite{PoT2022ABS221112588}. 

Since code generation is important in various code-related tasks and many reasoning tasks, many previous studies focus on achieving better code generation performance. 
Integrating feedback from the compiler is an intuitive way to help the model reflect on previous mistakes and generate better code.  
For instance, Self-Debug~\cite{SelfDebug2023ABS230405128} suggested that code LLMs be instructed to generate code, execute it, and subsequently improve the code quality based on its execution results. 
Additionally, Print-Debug~\cite{PrintDebugABS240105319} proposed to insert print statements to generate more detailed logs for debugging purposes. 
Furthermore, OpenCodeInterpreter~\cite{OpenCodeInterpreter2024ABS240214658} incorporated simulated human feedback into the interaction.
These studies have demonstrated that incorporating \seq sequences of code generation, execution, and analysis as CoTs can enhance the performance of code LLMs.

\begin{figure}[t]
    \centering
    \includegraphics[width=\columnwidth]{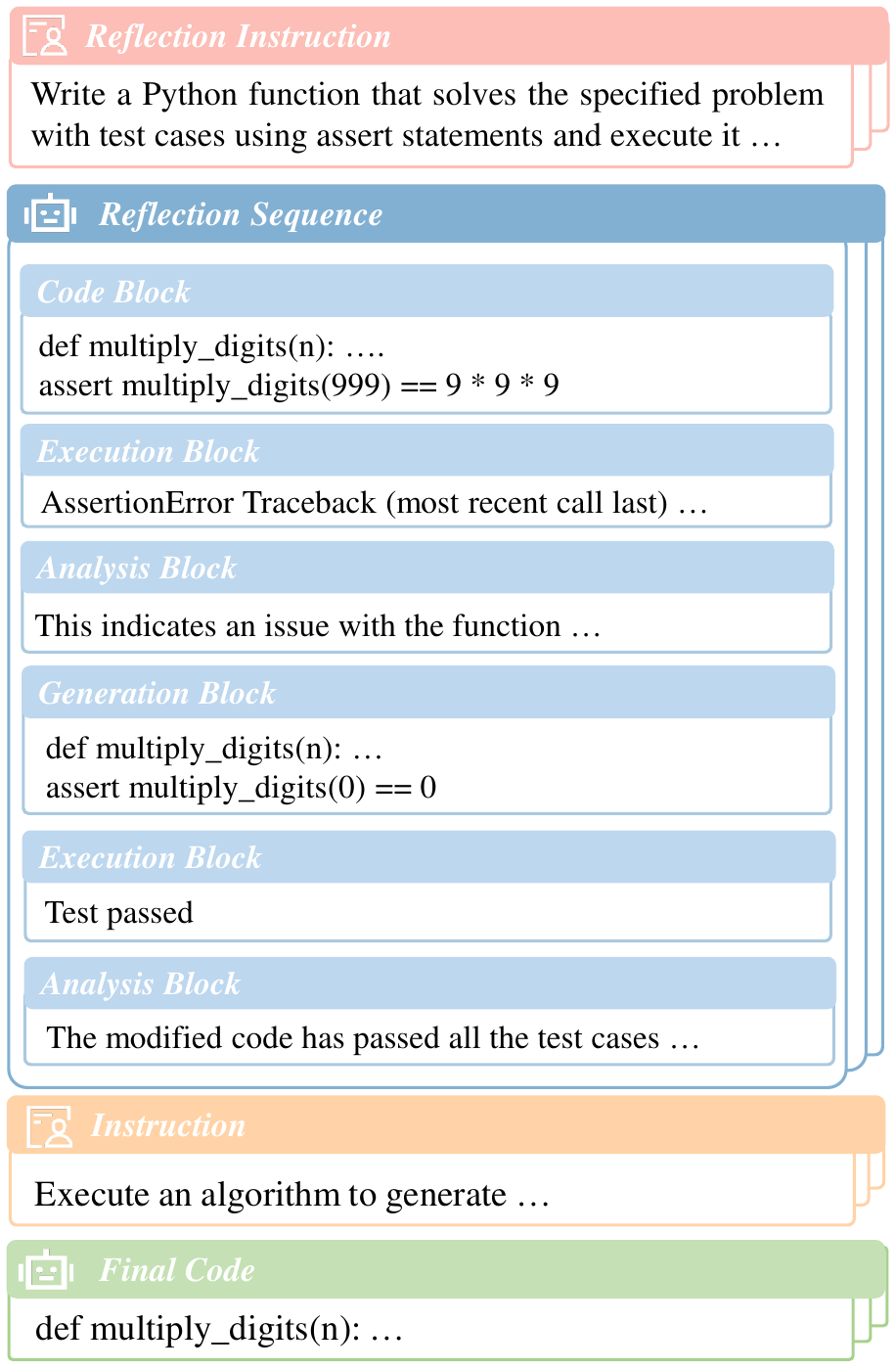}
    \caption{A sample of \seq sequence data containing four components: Reflection Instruction, Reflection Sequences, Instruction, and Final code.}
    \label{fig:method}
\end{figure}

Inspired by these works, we propose to leverage the \seq sequences to guide the fine-tuning of code LLMs. 
The proven effectiveness of \seq sequences as CoTs in enhancing the code generation performance demonstrates their inherent knowledge, which can guide model fine-tuning and result in better one-off code generation performance.
However, at least two challenges must be considered when using the \seq sequences to guide the model fine-tuning.
Firstly, the \seq sequences differ from the vanilla one-off code generation. Most of the codes in the \seq sequences are partly modified based on previous codes, while all codes are completed in the inference stage. The gap between the training and inference stages results in relatively low utilization of the \seq sequence.
Secondly, most of the codes in \seq sequence are generated based on previous executions and analysis, whereas a one-off generation relies solely on a single instruction. This disparity makes it challenging to transition between such different prompts effectively.

Based on these concerns, we proposed \name, a novel method to effectively leverage \seq sequence to perform better in one-off code generation tasks. 
To bridge the gap between the \seq sequences and the vanilla code generation, we propose \emph{\seq self-distillation}. Specifically, we carefully design a two-stage prompt to obtain high-quality instruction answer pairs with the same format as one-off generations. 
We first employ an LLM to generate a \seq sequence for an instruction with a compiler, and then task it to re-answer the instruction based on this sequence. After that, as shown in Figure~\ref{fig:method}, we obtain two rounds of dialogue as [Reflection Instruction, Reflection Sequence, Instruction, Final code]. The second round dialogue is the same as the one-off generation but with higher quality, which can play the role of a teacher sample distilling knowledge into one-off code generation.
To effectively distill knowledge from \seq sequence to one-off generation, we design a novel distillation method, namely \emph{dynamically masked distillation}. 
Specifically, with a particular LLM, the teacher input is the entire two-round dialogue, while the student input is a partly masked first-round dialogue along with an intact second-round dialogue.
During the training process, we gradually increase the masking rate to progressively enhance the difficulty of generating the final code.
In this way, LLM can be distilled to generate the final code from easy to difficult and achieve better performance.

Our contributions are summarized as follows:
\begin{itemize}
    \item We propose to leverage \seq sequences to improve the one-off code generation performance of code LLMs, which can be generated by LLMs and thus save annotation costs.
    \item On top of the idea, we propose two techniques, namely \emph{\seq self-distillation} and \emph{dynamically masked distillation}, which can effectively utilize the reflection sequence to improve the one-off code generation performance.
    \item Extensive experiments on HumanEval~(+), MBPP~(+), MultiPl-E, APPs, LiveCodeBench, ClassEval, and BigCodeBench demonstrate the effectiveness of the proposed method on one-off code generation. Notably, \name-DeepSeek-Coder-33B reaches 82.9~(76.8) on HumanEval~(+) and 84.1~(72.0) on MBPP~(+), which is an on-par performance of Claude-3-opus and surpasses early GPT-4.
\end{itemize}

%% file: sec-rel.tex
\section{Related Work} \label{sec:rel}

\subsection{Large Language Models for Code}
Large Language Models~\cite{InstuctGPT2022Ouyang0JAWMZASR22, GPT42023ABS230308774, Palm2023ABS230510403, LLama2023ABS230213971, LLama22023ABS230709288, Falcon2023PenedoMHCACPAL23, BaiChuan2023ABS230910305, Qwen2023ABS230916609, Mistral2023ABS231006825, Mixtral2023ABS240104088, Gemini2023ABS231211805, Claude2023Anthropic} have proven highly effective in general natural language processing~(NLP) tasks. 
For a specific domain such as code-related tasks~\cite{CodexABS210703374, MBPP2021ABS210807732, FIM2022ABS220714255, OctoPack2023ABS230807124}, training on large specific domain datasets can greatly improve their efficacy. 
Recent studies have introduced several LLMs for the code domain. 
OpenAI introduced Codex~\cite{CodexABS210703374}, and Google introduced PaLM-Coder~\cite{PaLM2023ChowdheryNDBMRBCSGSSTMRBTSPRDHPBAI23}. 
However, these models are closed-source, and we can only access them via API without access to their parameters.
There are also several open-source LLMs for the code domain, such as CodeGen~\cite{CodeGen2023NijkampPHTWZSX23}, Incoder~\cite{InCoderFriedAL0WSZYZL23}, SantaCoder~\cite{SantaCoderABS230103988}, StarCoder~\cite{StarCoderABS230506161}, StarCoder-2~\cite{StarCoder22024ABS240219173}, CodeGeeX~\cite{CodeGeeX2023ZhengXZDWXSW0LS23}, Code Llama~\cite{CodeLlamaABS230812950}, and DeepSeek-Coder~\cite{DeepSeekCoderABS240114196}. 
In addition to vanilla code snippets, modification content of code with commit messages~\cite{OctoPack2023ABS230807124} and code structure~\cite{ASTT52024ABS240103003} are also proposed to be the pre-train corpus. 
After instruction tuning, some of these open-source models have outperformed several closed-source models~\cite{WizardCoder2023ABS230608568}.

\subsection{Instruction Tuning for Code}
The primary objective of instruction tuning is training LLMs to align with human instructions by using a large corpus of human instructions together with corresponding responses~\cite{MultiTaskPrompt2022SanhWRBSACSRDBX22, SelfInstruct2022WeiBZGYLDDL22, InstuctGPT2022Ouyang0JAWMZASR22, Flan2023LongpreHVWCTZLZ23, Survey2023Ziyin}. 
Fine-tuning upon this method, LLMs can directly follow user instructions without extra demonstration and improve their generalization capacity. 
Its great value is also demonstrated in code-related applications. 
For example, Code Alpaca~\cite{CodeAlpaca2023GitHub} applied SELF-INSTRUCT~\cite{SelfInstruct2022WeiBZGYLDDL22} to fine-tune LLMs with ChatGPT-generated instructions. 
WizardCoder~\cite{WizardCoder2023ABS230608568} proposed Code Evol-Instruct, which evolves Code Alpaca data using the ChatGPT to generate more complex and diverse datasets.
PanGu-Coder2~\cite{PanGuCoder22023ABS230714936} proposed Rank Responses to align Test\&Teacher Feedback framework, which uses ranking responses as feedback instead of the absolute value of a reward model.
In addition to starting with instructions, a lot of work starts with existing source code. For example, MagiCoder~\cite{Magicoder2023ABS231202120}, WaveCoder~\cite{WaveCoderABS231214187}, and InverseCoder~\cite{InverseCoder2024Yutong} proposed some methods to make full use of source code. 

\subsection{Iterative Generation and Refinement}
Iterative refinement approaches are often taken to improve the generation quality. Recently, Self-Refine~\cite{SelfRefine2023MadaanTGHGW0DPY23} and Reflexion~\cite{Reflexion2023ShinnCGNY23} demonstrated that LLMs can reflect on previous generations, generate feedback, and give better generations based on feedback.
In the code domain, several tools can provide feedback for generated code, such as compiler, and other static tools. 
Integrating feedback from these tools can help the LLMs better reflect on themselves and generate better codes. For example, Self-Debugging~\cite{SelfDebug2023ABS230405128} and Print-Debugging~\cite{PrintDebugABS240105319} proposed to integrate the execution result of the code as a feedback message to obtain better performance. 
StepCoder~\cite{StepCoder2024ABS240201391} and OpenCodeInterpreter~\cite{OpenCodeInterpreter2024ABS240214658} involved executing and iteratively refining code as multi-turn interactions into instruction tuning, improving the model's debugging ability.  
Concurrently, AutoCoder~\cite{AutoCoder2024Bin} employed multi-turn interaction to obtain high-quality instruction data and then improve the one-off generation performance. In contrast, our method method introduces the reflection sequence into the training stage instead of just using it to filter the data.

%% file: sec-met.tex
\section{Methodology} \label{sec:method}

In this section, we present the methodological details of the proposed \name. We begin with a vanilla distillation, followed by a carefully designed method that comprehensively extracts knowledge from the \seq sequences and guides the model training.

\begin{figure*}[t]
    \centering
    \includegraphics[width=\textwidth]{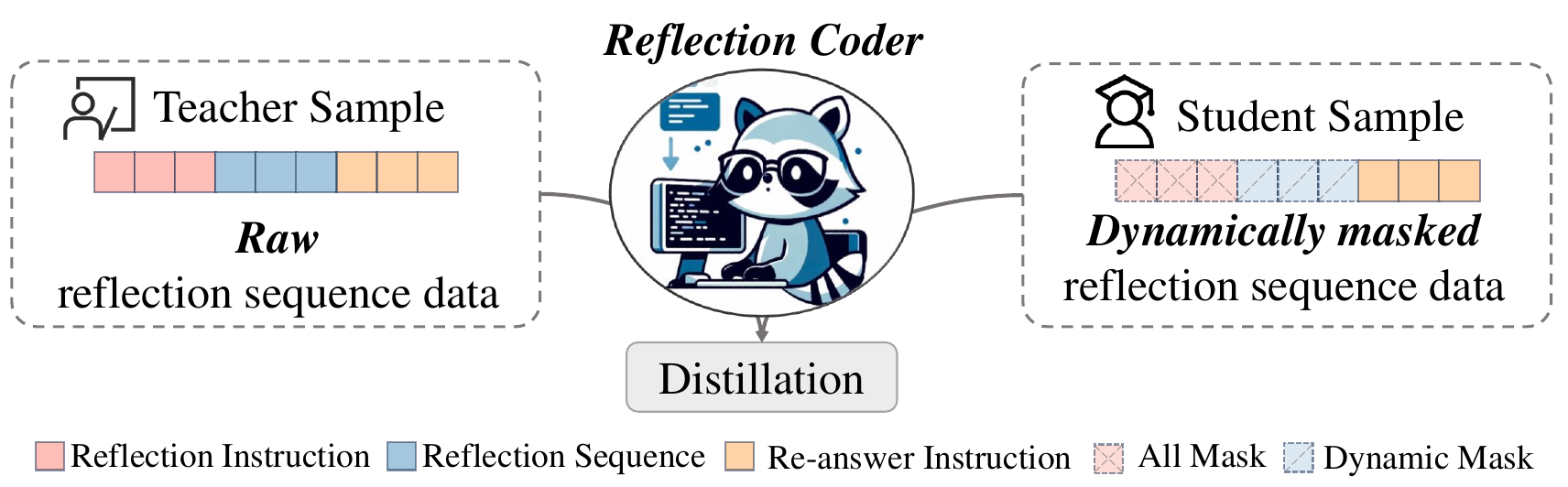}
    \caption{Overview of the proposed dynamically masked distillation.}
    \label{fig:distillation}
\end{figure*}

\subsection{Reflection Self-Distillation}

Here, we present how to utilize the \seq sequences to enhance the fine-tuning of code LLMs. As presented in Section~\ref{sec:intro}, a piece of \seq sequence data includes four components: [Reflection Instruction, Reflection Sequence, Instruction, Final code], where the \seq sequence is divided into three types of blocks, namely code block, execution block, and analysis block. Their contents are the generated executable code, the execution results, and the code summary or error analysis, respectively.

We construct two input samples for each \seq sequence to perform the \seq self-distillation. The teacher sample is the entire reflection sequence, and the student sample consists of [Instruction, Final Code], which is the same as vanilla one-off code generation instruction tuning data.
The key distinction between them is that the final code of the teacher sample can be generated based on the \seq sequences with low perplexity, while the student sample can only be generated according to the instruction. The vanilla distillation loss can be formulated as
\begin{equation}
    \mathcal{L}_{d}^{s} = \text{KL} \left( p(t_{c}|t_{ri},t_{rs},t_{i}) \parallel p(t_{c}|t_{i}) \right),
\end{equation}
where $t_{c}$ denotes tokens of the final code, $t_{ri}$ denotes tokens of the \seq instruction, $t_{rs}$ denotes tokens of the \seq sequence, and $t_{i}$ denotes tokens of the instruction. 

This approach enables the distillation of knowledge from the sequence into a one-off generation. The absolute position of the tokens in [Instruction, Final Code] differs between the teacher sample and the student sample, while [Reflection Instruction, Reflection Sequence] exists in the teacher sample but not in the student sample. However, the relative positions between the two tokens in [Instruction, Final Code] are the same between the teacher sample and the student sample, which indicates that distillation is effective for models utilizing Rotary Position Embedding~\cite{RoFormer2024SuALPBL24}, such as Llama~\cite{LLama22023ABS230709288}. 

\subsection{Dynamically Masked Distillation} 

Although vanilla distillation can distill knowledge from \seq sequence to enhance the one-off code generation, it could be hindered by the negative impact of contextual differences. Previous studies on distillation show that a student model distilled from a teacher with more parameters performs worse than the one distilled from a smaller teacher with a smaller capacity~\cite{Assistant2020MirzadehFLLMG20}. This finding suggests that the difference between teacher and student should not be too large. 
However, a significant gap exists between our teacher-student sample pair, as the teacher sample contains the entire reflection sequence while the student sample has no access to the reflection procedure. This discrepancy could lead to the poor performance of vanilla distillation.

Inspired by Curriculum Learning~\cite{CurriculumLearning2009BengioLCW09}, we carefully design a \emph{dynamically masked distillation} method. The overall procedure is presented in Figure~\ref{fig:distillation}. The initial student sample is the same as the teacher sample. During the training process, we mask all tokens of the ``Reflection Instruction'' and a portion of tokens of the ``Reflection Sequence''. The number of masked tokens is gradually increased to progressively enhance the difficulty of generating the final code, thereby enabling the model to effectively learn the knowledge encoded in the \seq sequence. Then the distillation loss can be formulated as
\begin{equation}
    \mathcal{L}_{d} = \text{KL} \left( p(t_{c}|t_{ri},t_{rs},t_{i}) \parallel p(t_{c}|t_{prs},t_{i}) \right),
\end{equation}
where $t_{prs}$ denoted tokens partly masked \seq instruction.

\begin{figure*}[t]
    \centering
    \includegraphics[width=\textwidth]{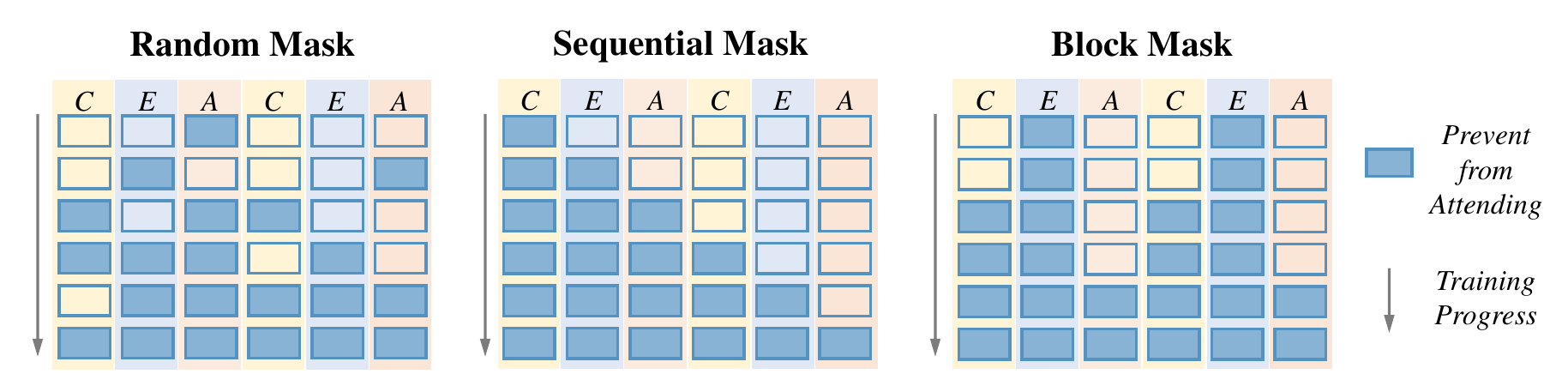}
    \caption{Overview of the proposed dynamic masking strategies. Here, a cell denotes a block, `C' denotes the code block, `E' denotes the execution block, and `A' denotes the analysis block.}
    \label{fig:mask}
\end{figure*}

As shown in Figure~\ref{fig:mask}, we design three dynamic masking strategies, namely \emph{random mask}, \emph{sequential mask}, and \emph{block mask}. All of these strategies adjust dynamically with the mask rate, a concept related to the training process, which can be defined as ``$\text{current step}/\text{max step}$''. The masking details are illustrated below:
\begin{enumerate}[leftmargin=5mm, itemsep=1mm, label=(\arabic*)]
    \item \emph{Random mask} selects blocks to mask based on the mask rate randomly. This is an intuitive strategy used by many previous studies in the pre-training stage such as BERT~\cite{BERT2019DevlinCLT19} and T5~\cite{T52020RaffelSRLNMZLL20}.
    \item \emph{Sequential mask} selects the leftmost blocks to mask and gradually expands the masked scope according to the mask rate. The underlying principle of this strategy is that later tokens are usually more influential in generating the final code since code generated after analysis tends to be more accurate than those generated initially.
    \item \emph{Block mask} selects some blocks according to mask rates. Specifically, when the mask rate exceeds $0$, all execution blocks are masked. When the mask rate exceeds $1/3$, all generation blocks are additionally masked. When the mask rate exceeds $2/3$, all analysis blocks are further masked. The core idea of this strategy is that the effectiveness of tokens is block-dependent. For instance, tokens in the execution block typically have the lowest impact.
\end{enumerate}
With these dynamically masked strategies, the learning difficulty gradually increases, contributing to better final one-off code generation performance. 
Similar to \emph{\seq self-distillation}, the absolute position of tokens in the one-off code generation round differs between the training stage and the inference stage, while ``Reflection Sequence'' exists in the training stage but not in the inference stage. However, the relative positions of the two tokens in [Instruction, Final Code] remain the same between the training stage and the inference stage, which indicates that there is no gap between the training stage and the inference stage for models utilizing Rotary Position Embedding~\cite{RoFormer2024SuALPBL24}. 

\paragraph{Training loss.} We employ both the next token prediction loss and distillation loss to train the model. For the teacher sample, we perform the next token prediction task on ``Final Code'' and the text blocks and the code blocks of ``Reflection Sequence'', because both the queries of the user and the execution results do not need to be generated in the inference stage. For the student sample, we only perform the next token prediction task on ``Final Code''.
The final loss consists of the next token prediction loss of the teacher and student samples, and the distillation loss between the teacher and student sample.

%% file: sec-exp.tex
\section{Experiments} \label{sec:exp}

\subsection{Experimental Setup} \label{sec:setup}

\begin{table*}[t]
\centering
\resizebox{\textwidth}{!}{
\begin{tabular}{l|c|cccc}
\toprule
\multirow{2.5}{*}{Method} & \multirow{2.5}{*}{Base} & \multicolumn{4}{c}{Benchmark} \\ \cmidrule(lr){3-6}
~ & ~ & HumanEval & HumanEval+ & MBPP & MBPP+ \\
\midrule \multicolumn{6}{c}{Closed-Source Models} \\ \midrule
O1-Preview~(Sept 2024)~\cite{O12024Aaron} & - & 96.3 & 89.0 & 95.5 & 80.2 \\
GPT-4-Turbo~(April 2024)~\cite{GPT42023ABS230308774} & - & 90.2 & 86.6 & - & - \\
GPT-4-Turbo~(Nov 2023)~\cite{GPT42023ABS230308774} & - & 88.4 & 81.7 & 85.7 & 73.3 \\
GPT-3.5-Turbo~(Nov 2023)~\cite{InstuctGPT2022Ouyang0JAWMZASR22} & - & 76.8 & 70.7 & 82.5 & 69.7 \\
Claude-3-opus~(Mar 2024)~\cite{Claude2023Anthropic} & - & 82.9 & 76.8 & 89.4 & 73.3 \\
Claude-3-sonnet~(Mar 2024)~\cite{Claude2023Anthropic} & - & 70.7 & 62.8 & 83.6 & 69.3 \\
Mistral Large (Mar 2024)~\cite{Mistral2023ABS231006825} & - & 70.1 & 62.8 & 72.8 & 59.5 \\
Gemini Pro 1.0~\cite{Gemini2023ABS231211805} & - & 63.4 & 55.5 & 75.4 & 61.4 \\
\midrule \multicolumn{6}{c}{Open-Source Models} \\ \midrule
WizardCoder~\cite{WizardCoder2023ABS230608568} & CL-7B & 48.2 & 40.9 & 58.5 & 49.5 \\ 
MagiCoder-S~\cite{Magicoder2023ABS231202120} & CL-7B & 70.7 & 66.5 & 70.6 & 60.1 \\ 
OpenCodeInterpreter~\cite{OpenCodeInterpreter2024ABS240214658} & CL-7B & 72.6 & 67.7 & 66.4 & 55.4 \\
\name & CL-7B & \textbf{75.0} & \textbf{68.9} & \textbf{72.2} & \textbf{61.4} \\ \midrule
WizardCoder~\cite{WizardCoder2023ABS230608568} & CL-34B & 73.2 & 64.6 & 75.1 & 63.2 \\ 
OpenCodeInterpreter~\cite{OpenCodeInterpreter2024ABS240214658} & CL-34B & 78.0 & 72.6 & 73.4 & 61.4 \\
Speechless~\cite{Speechless2023Huggingface} & CL-34B & 77.4 & 72.0 & 73.8 & 61.4 \\ 
\name & CL-34B & \textbf{78.0} & \textbf{73.8} & \textbf{80.2} & \textbf{67.5} \\ \midrule
DeepSeek-Coder-Instruct~\cite{DeepSeekCoderABS240114196} & DS-6.7B & 73.8 & 70.1 & 74.9 & 65.6 \\ 
MagiCoder-S~\cite{Magicoder2023ABS231202120} & DS-6.7B & 76.8 & 70.7 & 69.4 & 69.0 \\ 
OpenCodeInterpreter~\cite{OpenCodeInterpreter2024ABS240214658} & DS-6.7B & 77.4 & 73.8 & 76.5 & 66.4 \\
Artigenz-Coder~\cite{ArtigenzCoder2024Huggingface} & DS-6.7B & 75.6 & 72.6 & 80.7 & \textbf{69.6} \\
\name & DS-6.7B & \textbf{80.5} & \textbf{74.4} & \textbf{81.5} & \textbf{69.6} \\ \midrule
DeepSeek-Coder-Instruct~\cite{DeepSeekCoderABS240114196} & DS-33B & 81.1 & 75.0 & 80.4 & 70.1 \\ 
WizardCoder~\cite{WizardCoder2023ABS230608568} & DS-33B & 79.9 & 73.2 & 81.5 & 69.3 \\ 
OpenCodeInterpreter~\cite{OpenCodeInterpreter2024ABS240214658} & DS-33B & 79.3 & 73.8 & 80.2 & 68.5 \\
\name & DS-33B & \textbf{82.9} & \textbf{76.8} & \textbf{84.1} & \textbf{72.0} \\
\bottomrule 
\end{tabular}
}
\caption{Pass@1 accuracy on HumanEval(+) and MBPP(+). Here, `CL' denotes Code Llama, and `DS' denotes DeepSeek-Coder. The best results of each base are in bold and results unavailable are left blank.}
\label{tab:python}
\end{table*}

\paragraph{Training Dataset.} Our training dataset includes a vanilla code instruction tuning dataset, where each sample contains an instruction and corresponding code answer and the proposed reflection sequence dataset.
For the code instruction tuning dataset, we use instruction answer pairs from an open-source code instruction tuning dataset: CodeFeedback-Filtered-Instruction\footnote{\url{https://huggingface.co/datasets/m-a-p/CodeFeedback-Filtered-Instruction}}. 
For the reflection sequence dataset, we first randomly select 10k instructions with Python code in the corresponding answer to conduct two rounds of dialogue with GPT-4 Code Interpreter\footnote{\url{https://platform.openai.com/docs/assistants/tools/code-interpreter}}, obtaining the reflection sequence dataset. Subsequently, we use the 10k \seq sequence data and 156k code instruction tuning data to fine-tune DeepSeek-Coder 33B~\cite{DeepSeekCoderABS240114196}. Using this fine-tuned model, we generate additional 12k \seq sequence data. The detailed data construction process is presented in Appendix~\ref{sec:appendix-data}. Finally, we fine-tune the target model using 22k \seq sequence data and 156k code instruction tuning data.

\paragraph{Test Dataset.}
We evaluate our method on HumanEval~\cite{CodexABS210703374} and MBPP~\cite{MBPP2021ABS210807732}, two of the most widely used
benchmarks for code generation. Each task in these benchmarks includes a task description as the prompt
and a handful of test cases to check the correctness of the LLM-generated code.
Considering the insufficiency of test cases in these benchmarks, 
\citet{EvalPlus2023LiuXW023} proposed HumanEval+ and MBPP+, which contain 80×/35× more tests. Following prior work~\cite{EvalPlus2023LiuXW023, Magicoder2023ABS231202120, OpenCodeInterpreter2024ABS240214658}, we use greedy decoding to generate one sample and focus on comparing the pass@1 metric. Due to the limited space, we present evaluation experiments on more code-related benchmarks in Appendix~\ref{sec:appendix-exp}, including MultilPL-E~\cite{MultiPLE2022ABS220808227}, DS1000~\cite{DS10002023Lai0WZZZYFWY23}, APPs~\cite{APPs2021Dan}, LiveCodeBench~\cite{LiveCodeBench2024Naman}, ClassEval~\cite{ClassEval2023Xueying}, and BigCodeBench~\cite{BigCodeBench2024Terry}.

\paragraph{Implementation Details.} 
We test our methods on Code Llama Python 7B/34B and DeepSeek-Coder 6.7B/33B. We finetune all models for 2 epochs. We employ AdamW~\cite{AdamW2019LoshchilovH19} optimizer with a learning rate of 5e-5 for 6.7B/7B models and 2e-5 for 33B/34B models, a 0.05 warm-up ratio, and a cosine scheduler. We set the batch size as 512 and the max sequence length as 4096.
To efficiently train the computationally intensive models, we simultaneously employ DeepSpeed~\citep{ZERO2020RajbhandariRRH20} and Flash Attention~\citep{FlashAttention2023ABS230708691}. On 16 NVIDIA A800 80GB GPUs, the experiments on 7B models and 34B models take 3.5 hours and 25 hours, respectively.

In the training process, we up-sample 22k \seq sequence data by a factor of 2 and mix them with 156k code instruction tuning data. For samples in code instruction tuning data, we only employ the next token prediction as the training task, \aka we only calculate the causal language model loss. For samples in \seq sequence data, we use the proposed method to calculate the loss. We only use the block mask strategy in the order of execution block, analysis block, and code block. Although each strategy can bring benefits, mixing them is no longer beneficial in the experiments. 

\subsection{Evaluation}

\paragraph{Baselines.} We compare \name with previous state-of-the-art methods, including WizardCoder~\cite{WizardCoder2023ABS230608568}, Speechless~\cite{Speechless2023Huggingface}, DeepSeek-Coder Instruct~\cite{DeepSeekCoderABS240114196}, Magicoder~\cite{Magicoder2023ABS231202120}, and OpenCodeInterpreter~\cite{OpenCodeInterpreter2024ABS240214658}. All the results are consistently reported from the EvalPlus leaderboard\footnote{\url{https://evalplus.github.io/leaderboard.html}}. The proposed method is an instruction tuning method, so we do not present comparison results for base models such as StarCoder~\cite{StarCoderABS230506161} and Code Llama~\cite{CodeLlamaABS230812950}.

\paragraph{Results.} Table~\ref{tab:python} shows the pass@1 accuracy of different method on HumanEval~(+) and MBPP~(+). Based on the results, we have the following findings:
(1) For open-source methods with parameters ranging from 6.7B to 34B, the proposed \name outperforms previous state-of-the-art methods on all base models, demonstrating its effectiveness.
(2) Focusing on Code Llama, \name-CodeLlama-7B even surpasses WizardCode-CodeLlama-34B on HumanEval and HumanEval+.
(3) Compared with OpenCodeInterpreter, \name performs better on various base models, which indicates that we take better advantage of the reflection sequences.
(4) Compared with closed-source models, \name-DeepSeek-Coder-33B outperforms Gemini Pro, Mistral Large, and Claude-3-sonnet on all four benchmarks. It is worth noting that \name-DeepSeek-Coder-33B also achieves the on-par performance of GPT-3.5-Turbo and Claude-3-opus.

\subsection{Detailed Analysis}

Here, we conduct some analytical experiments. Due to the limited space, more analytical experiments are presented in the Appendix~\ref{sec:appendix-exp}.

\subsubsection{Ablation Study} \label{sec:ablation}

Here, we check how each component contributes to the final performance. We prepare three group variants of our method:
(1) The first group is related to the high-level method, which has three variants. \underline{w/o Dynamically Mask} denotes without any dynamically mask strategy, \aka the vanilla distillation. \underline{w/o Distillation} denotes without distillation, \aka only perform next token prediction on the \seq data. \underline{w/o Reflection Sequence} denotes without reflection sequence parts, \aka train the model on reflection data but without reflection sequences. \underline{w/o Reflection Data} denotes without \seq data, \aka only train the model with code instruction tuning data.
(2) The second group is related to the source of the \seq data. \underline{w/o GPT-4 Data} denotes only use the 12k \seq data construct from the fine-tuned DeepSeek-Coder 33B. Note that the DeepSeek-Coder 33B is fine-tuned with \seq Data from GPT-4. \underline{w/o DS Data} only use the 10k \seq data construct from GPT-4.
(3) The third group is related to the masking strategy. \underline{w/ Random Mask} and \underline{w/ Sequential Mask} denote replacing the block mask with random and sequential masks, respectively. \underline{w/ Three Mask Strategies} denotes randomly selecting a masking strategy in each step.

\begin{table}[t] \small
\centering
\setlength{\tabcolsep}{3pt}
\resizebox{\columnwidth}{!}{
\begin{tabular}{l|cc}
\toprule
Method & HumanEval~(+) & MBPP~(+) \\ \midrule
\name & \textbf{75.0~(68.9)} & \textbf{72.2~(61.4)} \\ \midrule
w/o Dynamic Mask & 70.7~(65.2) & 70.4~(58.5) \\
w/o Distillation & 69.5~(63.4) & 70.4~(59.0) \\
w/o Reflection Sequence & 66.5~(62.2) & 68.5~(57.9) \\
w/o Reflection Data & 65.9~(62.2) & 68.5~(57.9) \\ \midrule
w/o GPT Data  & 71.3~(67.1) & 70.1~(59.5) \\
w/o DS Data  & 68.9~(65.2) & 69.6~(58.2) \\ \midrule
w/ Random Mask & 72.0~(66.5) & 70.1~(59.0) \\
w/ Sequential Mask & 72.6~(67.7) & 71.3~(60.3) \\
w/ Three Strategies & 73.2~(65.9) & 71.7~(61.2) \\
\bottomrule 
\end{tabular}
}
\caption{Ablation results on HumanEval~(+) and MBPP~(+). The metric is Pass@1 accuracy, and all the results are based on Code Llama 7B.}
\label{tab:ablation}
\end{table}

Table~\ref{tab:ablation} shows the pass@1 accuracy of different variants on HumanEval~(+) and MBPP~(+). 
As we can see, the performance ranking can be given as: w/o Reflection Data < w/o Distillation < w/o Dynamically Mask < \name. These results indicate that all components are essential for improving performance. 
Moreover, w/o Reflection Sequence and w/o Reflection Data are almost the same. The main reason is that w/o Reflection Sequence are the same as the instruction tuning data in format, which does not introduce new knowledge into the training.
Additionally, both w/o GPT-4 Data and w/o DS Data perform worse than \name. And w/o GPT-4 Data performs better than w/o DS Data. A possible reason is that we have carried out strict filtering on Reflection Data from DS, which may impact the final performance.
Finally, w/ Random Mask, w/ Sequential Mask, and w/ Three Mask Strategies perform better than w/o Dynamically Mask but worse than \name. This indicates that while the three strategies are effective, they are not fully compatible with each other. A possible reason is that mixing them destroys the curricular nature of learning, leading to reduced effectiveness.

\begin{table}[t] 
\begin{tabular}{l|cc}
\toprule
Method & HumanEval~(+) & MBPP~(+) \\ \midrule
w/ EAC & \textbf{75.0~(68.9)} & \textbf{72.2~(61.4)} \\
w/ ECA & \textbf{75.0~(68.9)} & 70.9~(59.5) \\ 
w/ ACE & 72.0~(66.5) & 70.6~(60.1) \\ 
w/ AEC & 73.2~(65.9) & 70.9~(59.5) \\ 
w/ CAE & 71.3~(65.9) & 70.4~(59.8) \\ 
w/ CEA & 73.2~(67.1) & 72.0~(60.8) \\ 
\bottomrule 
\end{tabular}
\centering
\caption{Effect of masked order. The metric is Pass@1 accuracy, and all the results are based on Code Llama 7B. Here, `C' denotes the code block, `E' denotes the execution block, and `A' denotes the analysis block. For example, `ECA' denotes first mask execution block, then mask code block, and finally mask analysis block.}
\label{tab:order}
\end{table}

\subsubsection{Effect of Block Masked Order}

As mentioned in Section~\ref{sec:method}, the block mask masks block in a specific order. Here, we examine the effect of masking order by preparing six variants with all possible orders.

Table~\ref{tab:order} shows the pass@1 accuracy of different orders. 
As we can see, the two orders that mask execution blocks first perform better than other orders, indicating that tokens in execution blocks are generally less effective, which is intuitive. 
Similarly, the two orders that mask code blocks last also perform better, suggesting that tokens in code blocks are more effective.

\begin{table}[t]
\centering
\setlength{\tabcolsep}{3pt}
\resizebox{\columnwidth}{!}{
\begin{tabular}{c|ccc|cc}
\toprule
Model & GPT & 33B & 6.7B & HumanEval~(+) & MBPP~(+) \\ \midrule
33B & \Checkmark & \XSolidBrush & \XSolidBrush & 80.5~(73.8) & 80.7~(69.0) \\
33B & \Checkmark & \Checkmark & \XSolidBrush & \textbf{82.9~(76.8)} & \textbf{84.1~(72.0)} \\ \midrule
6.7B & \Checkmark & \Checkmark & \XSolidBrush & 80.5~(74.4) & \textbf{81.5~(69.6)} \\
6.7B & \Checkmark & \XSolidBrush & \Checkmark & 79.3~(76.2) & 80.7~(68.8) \\
6.7B & \XSolidBrush & \Checkmark & \XSolidBrush & 80.5~(75.0) & 81.0~(68.3) \\
6.7B & \XSolidBrush & \XSolidBrush & \Checkmark & \textbf{81.1~(76.2)} & 80.4~(68.3) \\
\bottomrule 
\end{tabular}
}
\caption{Effect of data source. The metric is Pass@1 accuracy. Here, ``33B'' denotes Deepseek-Coder-33B and ``6.7B'' denotes Deepseek-Coder-6.7B.}
\label{tab:data}
\end{table}

\subsubsection{Effect of Data Source}

As mentioned in Section~\ref{sec:setup}, our reflection sequence dataset is constructed from GPT-4 and fine-tuned Deepseek-Coder-33B. Here, we construct three sets of experiments to check the effectiveness of our method with different other data sources.

Firstly, we compared the ReflectionCoder-Deepseek-Coder-33B with the Deepseek-Coder-33B fine-tuned only with data from GPT-4, which is used to construct more data in our main experiments. As shown in the first group of Table~\ref{tab:data}, the intermediate model performs worse than the final model, which shows that the model can generate its training data and improve itself based on our method after only a small amount of training data from GPT-4. 
    
Then, we employ the Deepseek-Coder-6.7B to act as the intermediate model. As shown in the second group of Table~\ref{tab:data}, show that the data generated by the DeepSeek-Coder 6.7B can still bring benefits. Surprisingly, for HumanEval, the Deepseek-Coder 6.7B fine-tuned with self-generated reflection sequence data achieves better performance. The results also show that GPT-4 data is not the key to improving model performance. As long as the model learns how to reflect based on execution results, it can generate a reflection sequence for the model to improve itself.

\begin{table}[t]
\centering
\setlength{\tabcolsep}{3pt}
\resizebox{\columnwidth}{!}{
\begin{tabular}{l|cccc|cc}
\toprule
Model & HumanEval~(+) & MBPP~(+) \\ \midrule
Llama-3.1-8B-Instruct & 70.1~(62.2) & 72.5~(59.3) \\
~ w/ reflection & \textbf{76.2}~(64.7) & \textbf{74.2}~(62.2) \\
~ w/ distillation & 74.4~\textbf{(68.3)} & 73.0~\textbf{(63.0)} \\
~ w/ distillation \& reflection & 74.4~(67.7) & 72.2~(62.4) \\
\bottomrule 
\end{tabular}
}
\caption{Experiment on Llama-3.1-8B-Instruct. The metric is Pass@1 accuracy. Here, ``w/ reflection'' denotes performing reflection while testing on Llama-3.1-8B-Instruct. ``w/ distillation'' denotes the one-off generation performance of the model fine-tuned with self-generated reflection sequence data. ``w/ distillation \& reflection'' denotes performing reflection while testing on the model fine-tuned with the self-generated reflection sequence data.}
\label{tab:llama3}
\end{table}

\subsubsection{Autonomous Enhancement}

To completely exclude the factor of GPT-4, we employ an open-source model~(Llama-3.1-8B-Instruct~\cite{Llama312024Dubey}) that can generate reflection sequences without any training to act as the data source. 
We first employ reflection in testing as the reference, which first tasks the Llama-3.1-8B-Instruct to generate the reflection sequence and then tasks the model to generate the final code in the test stage. 
Then, we task the model to generate the reflection sequence data and use the generated data to fine-tune itself with the proposed method. 
    
As shown in Table~\ref{tab:llama3}, although Llama-3.1-8B-Instruct has undergone multiple rounds of post-training (including SFT and multi-turn DPO), our method can still further improve its performance and only rely on the data generated by itself. Surprisingly, on the plus dataset, the proposed method even outperforms w/ reflection and w/ distillation \& reflection. The reason is that the expected answers directly generated by the model have a high error rate, making it difficult to cover boundary data and more difficult data. Meanwhile, the data used for training has been strictly filtered (the filtered details are presented in Appendix~\ref{sec:appendix-data}), so the quality of the data used for training is relatively high. The filter pass rate~(only 17\%) also shows that the quality of the generated data is relatively low.

\subsection{Case Study}

Here, we present a case to show the effectiveness of the proposed \name. Specifically, the instruction is:
\begin{lstlisting}
def iscube(a):
    """
    Write a function that takes an integer a and returns True if this integer is a cube of some integer number. Note: you may assume the input is always valid.
    Examples:
    iscube(1) ==> True
    iscube(2) ==> False
    iscube(-1) ==> True
    iscube(64) ==> True
    iscube(0) ==> True
    iscube(180) ==> False
    """
\end{lstlisting}
We compare two generated implementations:
\begin{lstlisting}
# Response of "w/o Reflection Data"
def iscube(a):
    cube_root = round(a ** (1. / 3))
    return cube_root ** 3 == a
\end{lstlisting}
\begin{lstlisting}
# Response of ReflectionCoder
def iscube(a):
    if a < 0:
        a = -a
    cube_root = round(a ** (1. / 3.))
    return cube_root ** 3 == a
\end{lstlisting}

Although the solution generated by the ``w/o Reflection Data'' appears correct, it fails to account for the behavior of the ``round'' function when handling negative inputs. This oversight could lead to incorrect results for negative cube numbers. In contrast, the refined version—likely influenced by reflective feedback—correctly handles negative inputs by converting them to positive values before computing the cube root.

This case highlights a key advantage of leveraging reflection sequences: models can learn nuanced behaviors of library functions, such as ``round'', through feedback during the training process. Consequently, the model with access to reflection data demonstrates a deeper and more reliable understanding of function semantics.

\subsection{Generalization}

Here, we evaluate the generalization ability of our proposed methods. Specifically, we fine-tune a LLaMA-3.1-8B model using the training sets of GSM8K~\cite{GSM8k2021Karl} and MATH~\cite{Math2021Dan}. The baseline model is trained to directly predict the final answer without any chain-of-thought (CoT). For the proposed method, we regard the CoT as the reflection sequence and apply dynamic masking during distillation.

\begin{table}[t]
\centering
\begin{tabular}{lcc}
\toprule
\textbf{Model} & \textbf{GSM8K} & \textbf{MATH} \\
\midrule
w/o \name & 9.9  & 9.6  \\
w/ \name  & \textbf{11.1} & \textbf{13.6} \\
\bottomrule
\end{tabular}
\caption{Experiments on two mathematical reasoning datasets. The metric is Pass@1 accuracy.}
\label{tab:math_results}
\end{table}

As shown in Table~\ref{tab:math_results}, even with minimal adaptation, our method yields consistent improvements on both datasets. These results demonstrate the potential applicability of our approach in non-code reasoning tasks. Furthermore, our method opens up the possibility for dynamic compression of reasoning paths, which may significantly reduce latency.

%% file: sec-con.tex
\section{Conclusion} \label{sec:con}

In this paper, we proposed \name, a novel method to effectively leverage the reflection sequence constructed by integrating feedback from the compiler to achieve better one-off code generation performance. We proposed two training techniques to effectively utilize the \seq sequences data, namely \emph{\seq self-distillation} and \emph{dynamically masked distillation}. The \emph{\seq self-distillation} aims to distillation from \seq sequence to one-off code generation, and the \emph{dynamically masked distillation} aims to utilize the \seq sequence to achieve better performance effectively.
In the future, we plan to improve this method to dynamically reduce unnecessary reasoning paths for domains that need to show reasoning paths to simplify the model output.

\section*{Limitations} 

The primary limitation of this study is its reliance on a powerful model, such as the GPT-4 code interpreter, for constructing \seq sequence data. While this method ensures high precision and efficiency, it also incurs significant computational costs, which may limit its accessibility and scalability, particularly in resource-constrained environments. However, as large language models continue to evolve, open-source models like Llama 3.1 are beginning to exhibit similar capabilities. We anticipate that this limitation will diminish as these models become more advanced and widely available.
Furthermore, the reliance on Rotary Position Embedding introduces an additional restriction. While effective within the specific context of this study, it may limit the method's generalizability and adaptability to different architectures or alternative embedding strategies.

\section*{Ethics Statement}

The models utilized in this paper, StarCoder~\cite{StarCoderABS230506161}, Code Llama~\cite{CodeLlamaABS230812950}, Deepseek-Coder~\cite{DeepSeekCoderABS240114196} and Llama-3.1~\cite{Llama312024Dubey}, are licensed for academic research purposes. Furthermore, the data employed in this study, Code Instruction Tuning Dataset\footnote{\url{https://huggingface.co/datasets/m-a-p/CodeFeedback-Filtered-Instruction}}, is collected from Magicoder-OSS-Instruct\footnote{\url{https://huggingface.co/datasets/ise-uiuc/Magicoder-OSS-Instruct-75K}}, Python code subset of ShareGPT\footnote{\url{https://huggingface.co/datasets/ajibawa-2023/Python-Code-23k-ShareGPT}}, Magicoder-Evol-Instruct\footnote{\url{https://huggingface.co/datasets/ise-uiuc/Magicoder-Evol-Instruct-110K}}, and Evol-Instruct-Code\footnote{\url{https://huggingface.co/datasets/nickrosh/Evol-Instruct-Code-80k-v1}}. All of these datasets are constructed from GPT-3 or GPT-4, while OpenAI permit on research access\footnote{\url{https://openai.com/policies/}} and all theses datasets are licensed for research purposes.

\section*{Acknowledgment}
This study was supported in part by the Centre for Perceptual and Interactive Intelligence (CPII) Ltd., a CUHK-led InnoCentre under the InnoHK initiative of the Innovation and Technology Commission of the Hong Kong SAR Government, and in part by NSFC-RGC Project N\_CUHK498/24. Hongsheng Li is a PI of CPII under the InnoHK.

%% file: sec-app.tex
\clearpage\appendix\section*{Appendix}

\definecolor{mydarkred}{rgb}{0.7, 0, 0}
\definecolor{mydarkblue}{rgb}{0.51, 0.68, 0.82}
\definecolor{mylightblue}{rgb}{0.995, 1, 1}

\section{Data Construction} \label{sec:appendix-data}

As mentioned in Section~\ref{sec:setup}, our reflection sequence data is constructed from GPT-4 Code interpreter and fine-tuned Deepseek-Coder-33B. Here, we present details of data construction.

\subsection{GPT-4 Code Interpreter}

Previous studies~\cite{CSV2023ABS230807921, MathCoder2023ABS231003731} have revealed that GPT-4 Code Interpreter~\footnote{\url{https://platform.openai.com/docs/assistants/tools/code-interpreter}} can write and run Python code in a sandbox execution environment to solve challenging code and math problems.
It can iterate on the incorrect code it had previously generated by analyzing the cause of the failure and regenerating the code until it executes successfully. 
Based on its capability, we designed a two-stage method to prompt the GPT-4 Code Interpreter to construct the \seq sequence dataset.

In the first stage, we task the GPT-4 Code Interpreter to generate code to solve the given problem and test the code with assert statements. If the code fails any of these tests, the GPT-4 Code Interpreter will analyze the reasons for failure and regenerate the code with necessary corrections automatically. In this way, we get a reflection sequence of code generation, execution, and analysis, as presented in the blue blocks in Figure~\ref{fig:method}.
The prompt detail is shown below: 
\begin{tcolorbox}[breakable, colback=mylightblue, colframe=mydarkblue, fonttitle=\bfseries, title=The first round prompt]
Here is a programming problem for you to tackle: \\ \\
(1) Write a Python function that solves the specified problem with craft test cases using assert statements and execute it. Pay special attention to edge cases to thoroughly validate your solution's correctness. \\ \\
(2) If your code fails any of the tests, carefully examine the root cause of the failure. Make the necessary corrections to your code and then retest to confirm the fixes. \\ \\
Note: At this phase, your primary goal is to ensure the reliability of your code. There's no need to delve into in-depth problem analysis or strive for code optimization. \\ \\
\textbf{\# Programming Problem} \\
\textcolor{mydarkred}{\{problem\}}
\end{tcolorbox}

In the second stage, we task the GPT-4 Code Interpreter to generate the entire code based on the preceding \seq sequence. Additionally, we instruct the model to refrain from using any words related to the preceding \seq sequence, effectively simulating the one-off code generation.
In this way, we get the high-quality code answer, as presented in the green block in Figure~\ref{fig:method}.
The prompt detail is shown below: 
\begin{tcolorbox}[breakable, colback=mylightblue, colframe=mydarkblue, fonttitle=\bfseries, title=The second round prompt]
Then, your task is to create a precise solution for the given programming problem. \\ \\
Your answer should be complete and standalone, avoiding references to external resources or past exercises, and omit phrases such as "correct version". \\ \\
There is no requirement to execute the code or provide any test/usage example. Just present the code for the given problem between "\textasciigrave\textasciigrave\textasciigrave python" and "\textasciigrave\textasciigrave\textasciigrave ".
\end{tcolorbox}

\begin{table*}[h] \small
\centering
\begin{tabular}{l|c|cccccc}
\toprule
Model & Base & Java & JavaScript & C++ & PHP & Swift & Rust \\ \midrule
StarCoder & SC-15B & 28.5 & 31.7 & 30.6 & 26.8 & 16.7 & 24.5 \\
WizardCoder & SC-15B & 35.8 & 41.9 & 39.0 & 39.3 & 33.7 & 27.1 \\ \midrule
Code Llama-Python & CL-7B & 29.3 & 31.7 & 27.0 & 25.1 & 25.6 & 25.5 \\
MagiCoder & CL-7B & 36.4 & 45.9 & 36.5 & 39.5 & 33.4 & 30.6 \\
MagiCoder-S & CL-7B & 42.9 & 57.5 & 44.4 & 47.6 & 44.1 & 40.3 \\
\name  & CL-7B & \textbf{53.2} & \textbf{62.1} & \textbf{47.9} & \textbf{53.6} & \textbf{49.1} & \textbf{50.6} \\ \midrule
Code Llama-Python & CL-34B & 39.5 & 44.7 & 39.1 & 39.8 & 34.3 & 39.7 \\
WizardCoder & CL-34B & 44.9 & 55.3 & 47.2 & 47.2 & 44.3 & 46.2 \\
\name & CL-34B & \textbf{61.4} & \textbf{70.7} & \textbf{63.2} & \textbf{65.7} & \textbf{55.8} & \textbf{64.0} \\ \bottomrule 
\end{tabular}
\caption{Pass@1 accuracy results on MulitiPL-E. The best results of each base are in bold. Here, `SC' denotes StarCoder, and `CL' denotes Code Llama.}
\label{tab:multi}
\end{table*}

\begin{table*}[t] \small
\centering
\begin{tabular}{l|c|cccccccc}
\toprule
Model & Base & C++ & Java & PHP & TS & C\# & Bash & JavaScript \\ \midrule
DS Instruct & DS-6.7B   & 63.4 & \textbf{68.4} & \textbf{68.9} & 67.2 & \textbf{72.8} & 36.7 & \textbf{72.7}  \\
\name       & DS-6.7B   & \textbf{69.5} & 65.8 & 65.2 & \textbf{70.8} & 69.6 & \textbf{42.4} & 72.0 \\ \midrule
DS Instruct & DS-33B    & 68.9 & \textbf{73.4} & \textbf{72.7} & 67.9 & 74.1 & 43.0 & \textbf{73.9}  \\
\name       & DS-33B    & \textbf{70.8} & 70.9 & 72.0 & \textbf{72.3} & \textbf{74.7} & \textbf{45.6} & \textbf{73.9} \\ \bottomrule 
\end{tabular}
\caption{Pass@1 accuracy results on MulitiPL-E. The best results of each base are in bold. Here, `DS' denotes DeepSeek-Coder.}
\label{tab:multi-ds}
\end{table*}

\subsection{Deepseek-Coder-33B}

Due to the high cost of calling the GPT-4 Code Interpreter, we only construct 10k reflection sequence data using the prompt provided in Section~\ref{sec:method}. To generate more reflection sequence data, as described in Section~\ref{sec:exp}, 
We first fine-tune the DeepSeek-Coder 33B~\cite{DeepSeekCoderABS240114196} model using 10k reflection sequence data and 156k code instruction tuning data, which endows it with the capability to generate code and interpret feedback from the compiler. Then, we use this fine-tuned model to construct more reflection sequence data.

In the constructing stage, we randomly select another 70k instructions, whose corresponding answers contain Python code, to prompt the fine-tuned model. The following steps are performed to implement the reflection process.
\begin{itemize}[leftmargin=5mm]
    \item First, we prompt the fine-tuned model to generate a code block, which contains code and test samples. 
    
    \item Then, we employ a Jupyter Client to execute the code and concatenate the execution result to the prompt as an execution block.

    \item After that, the model generates an analysis block for the cause if the code sample fails any of the tests.

    \item The model will repeat the code generation and analyzing process until there is no error or it reaches a maximum of eight iterations.
\end{itemize}
We filter out 38k samples whose generated codes contain I/O operations that can be identified by keyword matching (\eg "open," "dump," "pip") or fail to resolve all errors within the maximum of eight iterations limitation.
After that, we filter out samples that only contain one iteration, \ie the first generated code passes all test cases, whose test samples may be too simple to ensure the correctness of the final code. In this stage, we filter out an additional 20k samples from the 32k samples generated in the previous stage and ultimately retain 12k high-quality samples.

To sum up, we first select 70k instructions to iteratively construct reflection data, where 38k samples are discarded as they contain I/O operations or exceed the maximum iteration limitation.
Finally, we filter out 20k samples with only one round of reflection, which may have some errors in the final code, and retain 12k high-quality samples.

\begin{table*}[t] \small
\centering
\begin{tabular}{l|c|cccccccc}
\toprule
Model & Base & plt & np & pd & py & scp & sk & tf & All \\ \midrule
Incoder & 6B & 28.3 & 4.4 & 3.1 & 4.4 & 2.8 & 2.8 & 3.8 & 7.4 \\ 
CodeGen-Mono & 16B & 31.7 & 10.9 & 3.4 & 7.0 & 9.0 & 10.8 & 15.2 & 11.7 \\ 
Code-Cushman-001 & - & 40.7 & 21.8 & 7.9 & 12.4 & 11.3 & 18.0 & 12.2 & 18.1 \\ \midrule
StarCoder & SC-15B & 51.7 & 29.7 & 11.4 & 21.4 & 20.2 & 29.5 & 24.5 & 26.0 \\
WizardCoder & SC-15B & 55.2 & 33.6 & 16.7 & 26.2 & 24.2 & 24.9 & 26.7 & 29.2 \\ \midrule
Code LLama & CL-7B & 55.3 & 34.5 & 16.4 & 19.9 & 22.3 & 17.6 & 28.5 & 28.0 \\
WizardCoder & CL-7B & 53.5 & 34.4 & 15.2 & 25.7 & 21.0 & 24.5 & 28.9 & 28.4 \\
MagiCoder & CL-7B & 54.6 & 34.8 & 19.0 & 24.7 & 25.0 & 22.6 & 28.9 & 29.9 \\
MagiCoder-S & CL-7B & 55.9 & 40.6 & \textbf{28.4} & 40.4 & \textbf{28.8} & 35.8 & \textbf{37.6} & 37.5 \\ \midrule
\name & CL-7B & \textbf{56.2} & \textbf{43.1} & 24.5 & \textbf{46.7} & 23.1 & 45.5 & 35.6 & \textbf{37.8} \\
~~w/o Relfexion Data & CL-7B & 56.0 & 42.7 & 23.0 & 43.6 & 26.7 & \textbf{45.8} & 35.6 & 37.4 \\
\bottomrule 
\end{tabular}
\caption{Pass@1 accuracy results on DS-1000~(Completion format). The best results of each base are in bold. Here, `SC' denotes StarCoder, `CL' denotes Code Llama.}
\label{tab:ds1000}
\end{table*}

\begin{table*}[t] \small
\centering
\begin{tabular}{l|c|c|cc|c}
\toprule
\multirow{2.5}{*}{Method} & \multirow{2.5}{*}{APPs} & \multirow{2.5}{*}{LiveCodeBench} & \multicolumn{2}{|c|}{ClassEval} & \multirow{2.5}{*}{BigCodeBench} \\ \cmidrule(lr){4-5}
~ & ~ & ~ & Class Level & Func Level & ~ \\ \midrule
MagiCoderS-DS-6.7B & 12.8 & 17.6 & 20.0 & 43.4 & 47.6 \\
OpenCodeInterpreter-DS-6.7B & 11.5 & 17.6 & 19.0 & 42.6 & 44.6  \\
ReflectionCoder-DS-6.7B & \textbf{14.1} & \textbf{18.4} & \textbf{25.0} & \textbf{44.0} & \textbf{47.9} \\ \midrule
OpenCodeInterpreter-DS-33B & 17.5 & 22.3 & 26.0 & 43.4 & 51.0 \\
ReflectionCoder-DS-33B & \textbf{20.2} & \textbf{22.7} & \textbf{28.0} & \textbf{50.4} & \textbf{52.9} \\
\bottomrule 
\end{tabular}
\caption{Pass@1 accuracy on APPs, LiveCodeBench, ClassEval, and BigCodeBench.}
\label{tab:other}
\end{table*}

\section{Additional Experiments} \label{sec:appendix-exp}

\subsection{MultiPL-E}

Following MagiCoder~\cite{Magicoder2023ABS231202120}, we evaluate six wide languages, \ie Java, JavaScirpt, C++, PHP, Swift, and Rust, using MultiPL-E~\cite{MultiPLE2022ABS220808227} benchmark. We employ StarCoder~\cite{StarCoderABS230506161}, WizardCoder~\cite{WizardCoder2023ABS230608568}, Code Llama~\cite{CodeLlamaABS230812950}, and MagiCoder~\cite{Magicoder2023ABS231202120} as baselines. For this comparison, we follow MagiCoder and WizardCoder to set temperature = 0.2, top\_p = 0.95, max\_length = 512, and num\_samples = 50. 
As shown in Table~\ref{tab:multi}, the proposed \name outperforms the previous state-of-the-art methods on both Code Llama 7B and Code Llama 34B. It shows that \seq sequence in Python is also helpful to other languages. Surprisingly, \name Code Llama 7B even surpassed WizardCoder Code Llama 34B, which further demonstrates the effectiveness of the proposed method.

In addition, we compare our method to DeepSeek-Coder Instruct~\cite{DeepSeekCoderABS240114196} on seven languages, which are reported in the DeepSeek-Coder paper, \ie C++, Java, PHP, TS, C\#, Bash, and JavaScript. For this comparison, we adopted a greedy search approach following the DeepSeek-Coder Instruct. 
As shown in Table~\ref{tab:multi-ds}, the proposed \name outperforms DeepSeek-Coder Instruct in most languages. Note that the DeepSeek-Coder Instruct is trained with 2B tokens, while our models are trained with 300M tokens, which also shows the effectiveness of our methods.
Our method outperforms DeepSeek-Coder Instruct in three languages on DeepSeek-Coder-6.7B and five languages on DeepSeek-Coder-33B, which shows that the larger model has a greater transfer ability.

\subsection{DS-1000}

We also evaluate our method on the DS-1000 dataset~\cite{DS10002023Lai0WZZZYFWY23}, which contains 1K distinct data science coding issues, ranging from 7 popular Python data science libraries. We employ Incoder~\cite{InCoderFriedAL0WSZYZL23}, CodeGen~\cite{CodeGen2023NijkampPHTWZSX23}, StarCoder~\cite{StarCoderABS230506161}, WizardCoder~\cite{WizardCoder2023ABS230608568}, Code Llama~\cite{CodeLlamaABS230812950}, and MagiCoder~\cite{Magicoder2023ABS231202120} as baselines. For this comparison, we follow MagiCoder to set temperature = 0.2, top\_p = 0.95, max\_length = 512, and num\_samples = 40. 

As shown in Table~\ref{tab:ds1000}, our model outperforms all baselines on average score. However, when comparing our method with and without Reflection Data, where the latter is trained exclusively with 156k one-off code generation data points, our method does not significantly improve the DS-1000 dataset. A key factor contributing to this outcome is the limited representation of data related to these seven libraries in our training set, primarily due to constraints in computational resources. For instance, the need for substantial GPU resources restricts our ability to fully leverage TensorFlow and PyTorch, while the requirement for multi-modal capabilities limits our utilization of Matplotlib. Despite these limitations, it is noteworthy that our method does not adversely affect the performance of tasks associated with these libraries.

\subsection{Other Test Set}

Here, we check the effectiveness of our method on more diverse tasks, such as APPs~\cite{APPs2021Dan} and LiveCodeBench~\cite{LiveCodeBench2024Naman}, ClassEval~\cite{ClassEval2023Xueying} and BigCodeBench~\cite{BigCodeBench2024Terry}. We construct experiments based on Deepseek-Coder-7B and Deepseek-Coder-33B. We employ MagiCoder~\cite{Magicoder2023ABS231202120} and OpenCodeInterpreter~\cite{OpenCodeInterpreter2024ABS240214658} as baselines, which used similar fine-tuning data as our models. We use greedy sampling to obtain the results in a zero-shot setting for both baselines and our method. Note that for LiveCodeBench, we report the result after 2023-09-01, which is the release date of Deepseek-Coder.

As shown in Table~\ref{tab:other},  our proposed method improves model accuracy on the four datasets, although there are no relative instructions in the training data. The results show that our method has better generalization. 

\begin{figure}[t] \small
    \centering
    \subfigure[HumanEval (+).]{ \label{fig:upsample-humaneval}
        \includegraphics[width=0.9\columnwidth]{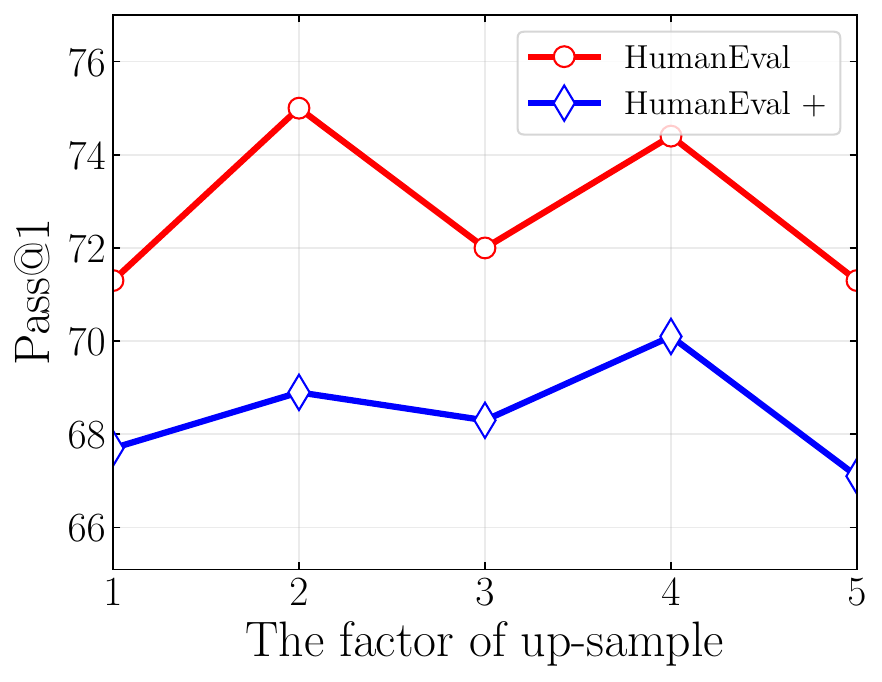}
    }
    \subfigure[MBPP (+).]{ \label{fig:upsample-mbpp}
        \includegraphics[width=0.9\columnwidth]{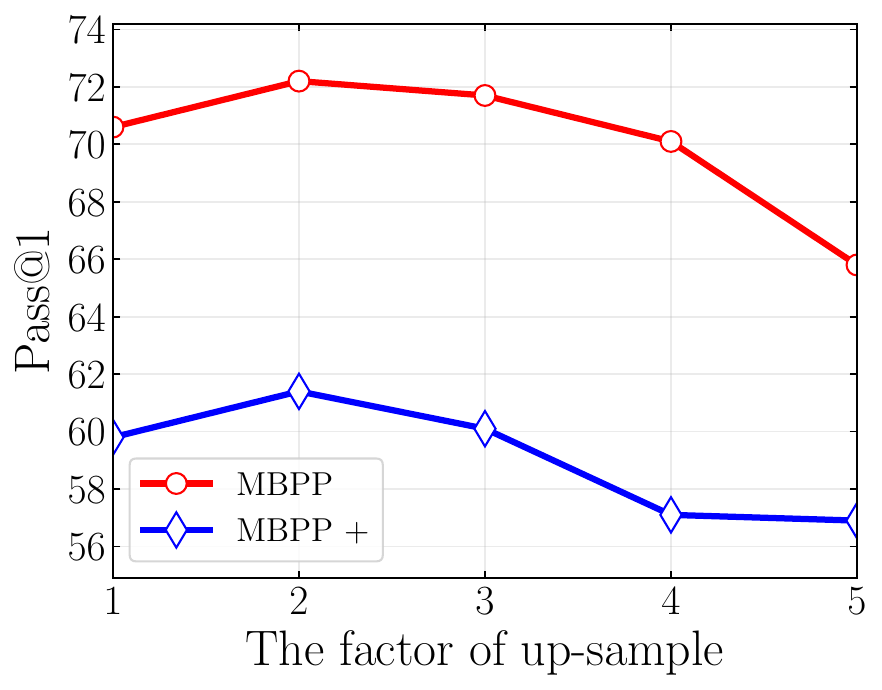}
    }
    \caption{Effect of the factor of up-sample. The metric is Pass@1 accuracy, and all the results are based on Code Llama 7B.}
\end{figure}

\begin{table}[t] \small
\centering
\begin{tabular}{l|cc}
\toprule
Method & HumanEval~(+) & MBPP~(+) \\ \midrule
\multicolumn{3}{c}{Code Llama 7B} \\ \midrule
\name & \textbf{75.0~(68.9)} & \textbf{72.2~(61.4)} \\ 
~~~w/o Reflection Data & 65.9~(62.2) & 68.5~(57.9) \\ \midrule
\multicolumn{3}{c}{Star Coder 7B} \\ \midrule
\name & \textbf{68.3~(63.4)} & 64.3~\textbf{(55.6)} \\ 
~~~w/o Reflection Data & 67.7~(62.8) & \textbf{66.7}~(54.8) \\ \bottomrule 
\end{tabular}
\caption{Effect of Rotary Position Embedding. The metric is Pass@1 accuracy.}
\label{tab:rope}
\end{table}

\subsection{Effect of the Factor of Up-sample}

As mentioned in Section~\ref{sec:exp}, we up-sample the \seq data and mix it with the code instruction tuning data. Here, we examine the effect of the up-sampling factor. Specifically, we vary the factor in the set \{1, 2, 3, 4, 5\}.
As shown in Figures~\ref{fig:upsample-humaneval} and \ref{fig:upsample-mbpp}, a factor of 2 results in optimal performance for most benchmarks. 
Due to the limited samples in HumanEval, the pass@1 fluctuates significantly. While a factor of 4 is optimal for HumanEval+, a factor of 2 remains optimal for HumanEval. 
A possible reason is that when the factor is too large, the \seq sequence data is repeated excessively, leading to overfitting and a consequent decrease in performance.

\subsection{Effect of Rotary Position Embedding}

As mentioned in Section~\ref{sec:method}, our method is effective for models utilizing Rotary Position Embedding because the absolute positions of the tokens of the answers in the teacher sample and the student sample are different, but the relative positions remain the same. Here, we construct an experiment to check the effect of Rotary Position Embedding on our method. Specifically, we perform our method and w/o Reflection Data on StarCoder, which uses an Absolute Position Embedding.

Table~\ref{tab:rope} shows the results on both Code Llama 7B~(w/ Rotary Position Embedding) and StarCoder 15B~(w/ Absolute Position Embedding).
As shown in the table, our method can effectively improve the performance of Code Llama 7B, but it is not so effective for StarCoder 15B. The primary reason is that the absolute positions of the tokens of the final answers are different for the training stage and the inference stage, which results in the distillation being biased.

\begin{table}[t] \small
\centering
\begin{tabular}{l|cc}
\toprule
Method & HumanEval~(+) & MBPP~(+) \\ \midrule
Random Mask & \textbf{72.0~(66.5)} & \textbf{70.1~(59.0)} \\
~~~w/ Token Level & 71.3~(66.5) & 68.8~(58.2) \\ \midrule
Sequential Mask & \textbf{72.6~(67.7)} & \textbf{71.3~(60.3)} \\
~~~w/ Token Level & 71.3~(67.1) & 68.5~(59.0) \\ \bottomrule 
\end{tabular}
\caption{Compare block-level mask strategies and token-level mask strategies. The metric is Pass@1 accuracy, and all the results are based on Code Llama 7B.}
\label{tab:token}
\end{table}

\subsection{Token-level Dynamic Masking Strategy}

In Section~\ref{sec:method}, we proposed three block-level dynamic masking strategies, namely random mask, sequential mask, and block mask. Here, we test our method with another two token-level dynamic masking strategies:
\begin{enumerate}[leftmargin=5mm, itemsep=1mm, label=(\arabic*)]
    \item \emph{Random Token mask} selects tokens to mask based on the mask rate randomly.
    \item \emph{Sequential Token mask} selects the leftmost tokens to mask and gradually expands the masked scope according to the mask rate. 
\end{enumerate}

Table~\ref{tab:token} shows the results on both block-level masking strategies and token-level masking strategies. The block-level masking strategies significantly outperform token-level masking strategies. Because the token-level masking strategies may destroy the integrity of texts or codes.

\begin{figure}[t] 
    \centering
    \includegraphics[width=0.9\columnwidth]{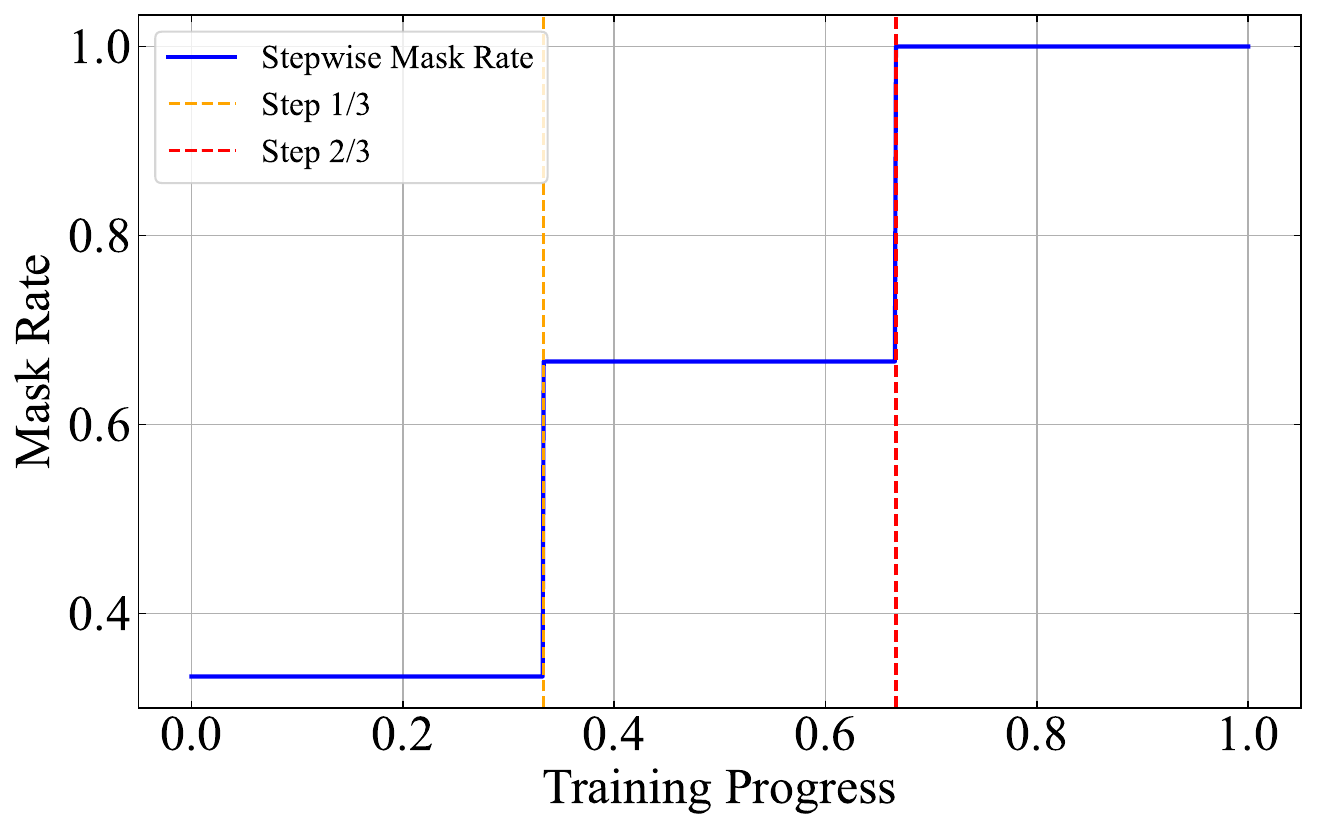}
    \caption{The changes in masked rate during training.}
    \label{fig:masked_rate}
\end{figure}

\begin{figure}[t] 
    \centering
    \includegraphics[width=\columnwidth]{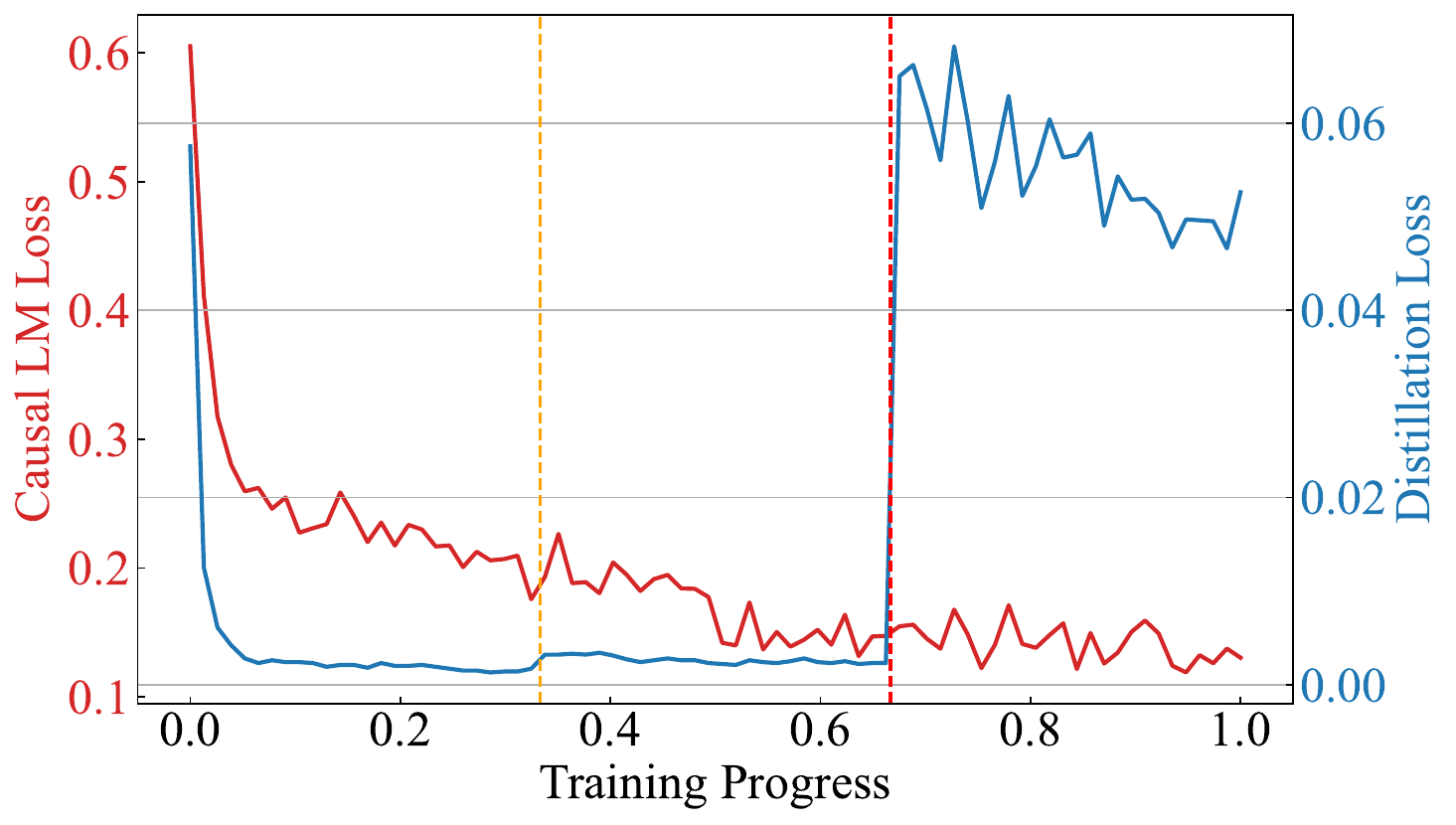}
    \caption{The changes in value of two loss components during training.}
    \label{fig:loss}
\end{figure}

\subsection{Training Procedure}

Here, we visualize the evolution of the masking rate and the training losses throughout the learning process. As shown in Figure~\ref{fig:masked_rate}, the masking schedule follows a stepwise curriculum strategy: during the first third of the training, all execution blocks are masked; in the second third, generation blocks are additionally masked; and in the final third, analysis blocks are further masked. This progressive masking design encourages the model to gradually adapt to more challenging conditions and enhances its robustness in code generation. As shown in Figure~\ref{fig:loss}, the causal language modeling loss steadily decreases over time, indicating consistent improvements in the model’s predictive ability. In contrast, the distillation loss displays sharp spikes at each one-third interval, followed by a gradual decline. These abrupt increases are caused by the sudden introduction of new masked block types, which temporarily raise the difficulty of the distillation task until the model re-stabilizes.

\section{Training Sample} \label{sec:appendix-sample}

Here, we present how to organize training samples. Following chat templates of most models, we use two special tokens to wrap the message of user and assistant, \ie ``<|user|>'', ``<|assistant|>'', and ``<|endofmessage|>''. Furthermore, we use two special tokens to wrap the content of different blocks, \ie  ``<|text|>'', ``<|code|>'', ``<|execution|>'', and ``<|endofblock|>''. Then, we concatenate all messages directly to form a piece of training data. Note that the user message and execution block are masked when calculating the loss because the environment provides the two contents during actual use.  Here is an example:

\begin{tcolorbox}[breakable, colback=background, colframe=title, fonttitle=\bfseries, title=Reflection Instruction]
\setstretch{1.125}
\textcolor{red}{<|user|>} \\
\textcolor{red}{<|text|>} \\
Here is a programming challenge for you to tackle: \\
(1) Write a Python function that solves the specified problem with craft test cases using assert statements and execute it. Pay special attention to edge cases to thoroughly validate your solution's correctness. \\
(2) If your code fails any of the tests, carefully examine the root cause of the failure. Make the necessary corrections to your code and then retest to confirm the fixes. \\ \\
\#\#\# Problem \\
Execute an algorithm to generate the numerical multiplication of individual integer components encapsulated within a larger numeric figure. \\
n = 145 \\
\textcolor{red}{<|endofblock|>} \\
\textcolor{red}{<|endofmessage|>}
\end{tcolorbox}
\begin{tcolorbox}[breakable, colback=background, colframe=title, fonttitle=\bfseries, title=Reflection Sequence]
\setstretch{1}
\textcolor{red}{<|assistant|>} \\
\textcolor{red}{<|code|>} \\
def multiply \_digits(n): \\
\text{~~~~}""" \\
\text{~~~~}This function takes an integer n and returns the product of its digits. \\
\text{~~~~}""" \\
\text{~~~~}product = 1 \\
\text{~~~~}while n > 0: \\
\text{~~~~~~~~}product *= n \% 10 \\
\text{~~~~~~~~}n //= 10 \\ \\
\text{~~~~}return product \\ \\
\# Test cases \\
assert multiply \_digits(145) == 4 * 5 * 1 \\
assert multiply \_digits(123) == 1 * 2 * 3 \\
assert multiply \_digits(999) == 9 * 9 * 9 \\
assert multiply \_digits(0) == 0 \\
assert multiply \_digits(10) == 1 * 0 \\
\textcolor{red}{<|endofblock|>} \\
\textcolor{red}{<|execution|>} \\
--------------------------------------------------------------------------- \\
AssertionError\text{~~~~~~~~~~~~~~~~~~~~~~~~~~~~}Traceback (most recent call last) \\
Cell In[1], line 15 \\
\text{~~~~~~}13 assert multiply \_digits(123) == 1 * 2 * 3 \\
\text{~~~~~~}14 assert multiply \_digits(999) == 9 * 9 * 9 \\
---> 15 assert multiply \_digits(0) == 0 \\
\text{~~~~~~}16 assert multiply \_digits(10) == 1 * 0 \\
 \\
AssertionError: \\
\textcolor{red}{<|endofblock|>} \\
\textcolor{red}{<|text|>} \\
The assertion `assert multiply \_digits(0) == 0` failed. This indicates an issue with the function when handling the case where `n` is 0. \\
 \\
In the current implementation, when `n` is 0, the while loop is never entered, and thus the function returns the initial value of `product`, which is 1. However, the correct behavior should be to return 0, as the product of the digits of 0 is 0. \\
 \\
I'll correct the function to handle this special case and retest the code. \\
\textcolor{red}{<|endofblock|>} \\
\textcolor{red}{<|code|>} \\
def multiply \_digits(n): \\
\text{~~~~}""" \\
\text{~~~~}This function takes an integer n and returns the product of its digits. \\
\text{~~~~}""" \\
\text{~~~~}if n == 0: \\
\text{~~~~~~~~}return 0 \\ \\
\text{~~~~}product = 1 \\
\text{~~~~}while n > 0: \\
\text{~~~~~~~~}product *= n \% 10 \\
\text{~~~~~~~~}n //= 10 \\ \\
\text{~~~~}return product \\ \\
\# Retesting with the original test cases and the corrected case \\
assert multiply \_digits(145) == 4 * 5 * 1 \\
assert multiply \_digits(123) == 1 * 2 * 3 \\
assert multiply \_digits(999) == 9 * 9 * 9 \\
assert multiply \_digits(0) == 0 \\
assert multiply \_digits(10) == 1 * 0 \\
\textcolor{red}{<|endofblock|>} \\
\textcolor{red}{<|execution|>} \\
\textcolor{red}{<|endofblock|>} \\
\textcolor{red}{<|text|>} \\
The modified code has passed all the test cases, including the previously failed case for `n = 0`. The function `multiply \_digits` now correctly handles all tested scenarios. \\
\textcolor{red}{<|endofblock|>} \\
\textcolor{red}{<|endofmessage|>}
\end{tcolorbox}
\begin{tcolorbox}[breakable, colback=background, colframe=title, fonttitle=\bfseries, title=Instruction]
\setstretch{1.25}
\textcolor{red}{<|user|>} \\
\textcolor{red}{<|text|>} \\
Execute an algorithm to generate the numerical multiplication of individual integer components encapsulated within a larger numeric figure. \\
n = 145 \\
\textcolor{red}{<|endofblock|>} \\
\textcolor{red}{<|endofmessage|>}
\end{tcolorbox}
\begin{tcolorbox}[breakable, colback=background, colframe=title, fonttitle=\bfseries, title=Final Answer]
\setstretch{1}
\textcolor{red}{<|assistant|>} \\
\textcolor{red}{<|text|>} \\
\textasciigrave\textasciigrave\textasciigrave python \\
def multiply \_digits(n): \\
\text{~~~~}""" \\
\text{~~~~}This function takes an integer n and returns the product of its digits. \\
\text{~~~~}""" \\
\text{~~~~}if n == 0: \\
\text{~~~~~~~~}return 0 \\
 \\
\text{~~~~}product = 1 \\
\text{~~~~}while n > 0: \\
\text{~~~~~~~~}product *= n \% 10 \\
\text{~~~~~~~~}n //= 10 \\ \\
\text{~~~~}return product \\
\textasciigrave\textasciigrave\textasciigrave \\
\textcolor{red}{<|endofblock|>} \\
\textcolor{red}{<|endofmessage|>}
\end{tcolorbox}

\section{Test Prompt} \label{sec:appendix-test}

Here, we present the test prompt for each test set to facilitate reproduction.

For HumanEval, we add ``Write a complete Python function for the problem'' before the original prompt from EvalPlus~\cite{EvalPlus2023LiuXW023} to make the model follow the instructions better.

\begin{tcolorbox}[breakable, colback=background, colframe=title, fonttitle=\bfseries, title=Test Prompt for HumanEval]
\setstretch{1.25}
<|user|><|text|>Write a complete Python function for the problem. \\ \\
\{prompt\}<|endofblock|><|endofmessage|>\\
<|assistant|><|text|>\textasciigrave\textasciigrave\textasciigrave python 
\end{tcolorbox}

For MBPP, we follow WizardCoder~\cite{WizardCoder2023ABS230608568} and Magicoder~\cite{Magicoder2023ABS231202120} to re-construct the source problem from EvalPlus~\cite{EvalPlus2023LiuXW023} with some additional prompts. 

\begin{tcolorbox}[breakable, colback=background, colframe=title, fonttitle=\bfseries, title=Test Prompt for MBPP]
\setstretch{1.25}
<|user|><|text|>
\{prompt\} \\
Your code should satisfy the following assertion: \\
\textasciigrave\textasciigrave\textasciigrave python \\
\{test sample\} \\
\textasciigrave\textasciigrave\textasciigrave <|endofblock|><|endofmessage|>\\
<|assistant|><|text|>\textasciigrave\textasciigrave\textasciigrave python 
\end{tcolorbox}

For MultiPL-E, we put the source prompts in both the user message and the beginning of the assistant message to ensure that the model does not modify the prompt and only completes it. 

\begin{tcolorbox}[breakable, colback=background, colframe=title, fonttitle=\bfseries, title=Test Prompt for MultiPL-E]
\setstretch{1.25}
<|user|><|text|>Write a complete \{language\} function for the problem. \\ \\
\{prompt\}<|endofblock|><|endofmessage|>\\
<|assistant|><|text|> \textasciigrave\textasciigrave\textasciigrave \{language\} \\
\{prompt\}
\end{tcolorbox}

For DS-1000, we directly use the source prompts.

For APPs and LiveCodeBench, we add ``Write a complete Python script for the question, Please note that you need to handle the stdin input, e.g. t = int(input()).'' before the original prompt to make the model follow the instructions better.

\begin{tcolorbox}[breakable, colback=background, colframe=title, fonttitle=\bfseries, title=Test Prompt for APPs / LiveCodeBench]
\setstretch{1.25}
<|user|><|text|>Write a complete Python script for the question, Please note that you need to handle the stdin input, e.g. t = int(input()). \\ \\
\{prompt\}<|endofblock|><|endofmessage|>\\
<|assistant|><|text|> \textasciigrave\textasciigrave\textasciigrave python
\end{tcolorbox}

For ClassEval, we add ``Please complete the class \{class name\} in the following code.'' before the original prompt to make the model follow the instructions better.

\begin{tcolorbox}[breakable, colback=background, colframe=title, fonttitle=\bfseries, title=Test Prompt for ClassEval]
\setstretch{1.25}
<|user|><|text|>Please complete the class \{class name\} in the following code. \\ \\
\{prompt\}<|endofblock|><|endofmessage|>\\
<|assistant|><|text|> \textasciigrave\textasciigrave\textasciigrave python
\end{tcolorbox}

%% file: main.bbl
\begin{thebibliography}{70}
\expandafter\ifx\csname natexlab\endcsname\relax\def\natexlab#1{#1}\fi

\bibitem[{Allal et~al.(2023)Allal, Li, Kocetkov, Mou, Akiki, Ferrandis, Muennighoff, Mishra, Gu, Dey, Umapathi, Anderson, Zi, Lamy{-}Poirier, Schoelkopf, Troshin, Abulkhanov, Romero, Lappert, Toni, del R{\'{\i}}o, Liu, Bose, Bhattacharyya, Zhuo, Yu, Villegas, Zocca, Mangrulkar, Lansky, Nguyen, Contractor, Villa, Li, Bahdanau, Jernite, Hughes, Fried, Guha, de~Vries, and von Werra}]{SantaCoderABS230103988}
Loubna~Ben Allal, Raymond Li, Denis Kocetkov, Chenghao Mou, Christopher Akiki, Carlos~Mu{\~{n}}oz Ferrandis, Niklas Muennighoff, Mayank Mishra, Alex Gu, Manan Dey, Logesh~Kumar Umapathi, Carolyn~Jane Anderson, Yangtian Zi, Joel Lamy{-}Poirier, Hailey Schoelkopf, Sergey Troshin, Dmitry Abulkhanov, Manuel Romero, Michael Lappert, Francesco~De Toni, Bernardo~Garc{\'{\i}}a del R{\'{\i}}o, Qian Liu, Shamik Bose, Urvashi Bhattacharyya, Terry~Yue Zhuo, Ian Yu, Paulo Villegas, Marco Zocca, Sourab Mangrulkar, David Lansky, Huu Nguyen, Danish Contractor, Luis Villa, Jia Li, Dzmitry Bahdanau, Yacine Jernite, Sean Hughes, Daniel Fried, Arjun Guha, Harm de~Vries, and Leandro von Werra. 2023.
\newblock \href {https://doi.org/10.48550/ARXIV.2301.03988} {Santacoder: don't reach for the stars!}
\newblock \emph{CoRR}, abs/2301.03988.

\bibitem[{Anil et~al.(2023{\natexlab{a}})Anil, Borgeaud, Wu, Alayrac, Yu, Soricut, Schalkwyk, Dai, Hauth, Millican, Silver, Petrov, Johnson, Antonoglou, Schrittwieser, Glaese, Chen, Pitler, Lillicrap, Lazaridou, Firat, Molloy, Isard, Barham, Hennigan, Lee, Viola, Reynolds, Xu, Doherty, Collins, Meyer, Rutherford, Moreira, Ayoub, Goel, Tucker, Piqueras, Krikun, Barr, Savinov, Danihelka, Roelofs, White, Andreassen, von Glehn, Yagati, Kazemi, Gonzalez, Khalman, Sygnowski, and et~al.}]{Gemini2023ABS231211805}
Rohan Anil, Sebastian Borgeaud, Yonghui Wu, Jean{-}Baptiste Alayrac, Jiahui Yu, Radu Soricut, Johan Schalkwyk, Andrew~M. Dai, Anja Hauth, Katie Millican, David Silver, Slav Petrov, Melvin Johnson, Ioannis Antonoglou, Julian Schrittwieser, Amelia Glaese, Jilin Chen, Emily Pitler, Timothy~P. Lillicrap, Angeliki Lazaridou, Orhan Firat, James Molloy, Michael Isard, Paul~Ronald Barham, Tom Hennigan, Benjamin Lee, Fabio Viola, Malcolm Reynolds, Yuanzhong Xu, Ryan Doherty, Eli Collins, Clemens Meyer, Eliza Rutherford, Erica Moreira, Kareem Ayoub, Megha Goel, George Tucker, Enrique Piqueras, Maxim Krikun, Iain Barr, Nikolay Savinov, Ivo Danihelka, Becca Roelofs, Ana{\"{\i}}s White, Anders Andreassen, Tamara von Glehn, Lakshman Yagati, Mehran Kazemi, Lucas Gonzalez, Misha Khalman, Jakub Sygnowski, and et~al. 2023{\natexlab{a}}.
\newblock \href {https://doi.org/10.48550/ARXIV.2312.11805} {Gemini: {A} family of highly capable multimodal models}.
\newblock \emph{CoRR}, abs/2312.11805.

\bibitem[{Anil et~al.(2023{\natexlab{b}})Anil, Dai, Firat, Johnson, Lepikhin, Passos, Shakeri, Taropa, Bailey, Chen, Chu, Clark, Shafey, Huang, Meier{-}Hellstern, Mishra, Moreira, Omernick, Robinson, Ruder, Tay, Xiao, Xu, Zhang, {\'{A}}brego, Ahn, Austin, Barham, Botha, Bradbury, Brahma, Brooks, Catasta, Cheng, Cherry, Choquette{-}Choo, Chowdhery, Crepy, Dave, Dehghani, Dev, Devlin, D{\'{\i}}az, Du, Dyer, Feinberg, Feng, Fienber, Freitag, Garcia, Gehrmann, Gonzalez, and et~al.}]{Palm2023ABS230510403}
Rohan Anil, Andrew~M. Dai, Orhan Firat, Melvin Johnson, Dmitry Lepikhin, Alexandre Passos, Siamak Shakeri, Emanuel Taropa, Paige Bailey, Zhifeng Chen, Eric Chu, Jonathan~H. Clark, Laurent~El Shafey, Yanping Huang, Kathy Meier{-}Hellstern, Gaurav Mishra, Erica Moreira, Mark Omernick, Kevin Robinson, Sebastian Ruder, Yi~Tay, Kefan Xiao, Yuanzhong Xu, Yujing Zhang, Gustavo~Hern{\'{a}}ndez {\'{A}}brego, Junwhan Ahn, Jacob Austin, Paul Barham, Jan~A. Botha, James Bradbury, Siddhartha Brahma, Kevin Brooks, Michele Catasta, Yong Cheng, Colin Cherry, Christopher~A. Choquette{-}Choo, Aakanksha Chowdhery, Cl{\'{e}}ment Crepy, Shachi Dave, Mostafa Dehghani, Sunipa Dev, Jacob Devlin, Mark D{\'{\i}}az, Nan Du, Ethan Dyer, Vladimir Feinberg, Fangxiaoyu Feng, Vlad Fienber, Markus Freitag, Xavier Garcia, Sebastian Gehrmann, Lucas Gonzalez, and et~al. 2023{\natexlab{b}}.
\newblock \href {https://doi.org/10.48550/ARXIV.2305.10403} {Palm 2 technical report}.
\newblock \emph{CoRR}, abs/2305.10403.

\bibitem[{Anthropic(2024)}]{Claude2023Anthropic}
Anthropic. 2024.
\newblock \href {https://www-cdn.anthropic.com/de8ba9b01c9ab7cbabf5c33b80b7bbc618857627/Model_Card_Claude_3.pdf} {The claude 3 model family: Opus, sonnet, haiku}.

\bibitem[{Artigenz-Coder(2024)}]{ArtigenzCoder2024Huggingface}
Artigenz-Coder. 2024.
\newblock \href {https://huggingface.co/Artigenz/Artigenz-Coder-DS-6.7B} {Artigenz-coder-ds-6.7b}.

\bibitem[{Austin et~al.(2021)Austin, Odena, Nye, Bosma, Michalewski, Dohan, Jiang, Cai, Terry, Le, and Sutton}]{MBPP2021ABS210807732}
Jacob Austin, Augustus Odena, Maxwell~I. Nye, Maarten Bosma, Henryk Michalewski, David Dohan, Ellen Jiang, Carrie~J. Cai, Michael Terry, Quoc~V. Le, and Charles Sutton. 2021.
\newblock \href {http://arxiv.org/abs/2108.07732} {Program synthesis with large language models}.
\newblock \emph{CoRR}, abs/2108.07732.

\bibitem[{Bai et~al.(2023)Bai, Bai, Chu, Cui, Dang, Deng, Fan, Ge, Han, Huang, Hui, Ji, Li, Lin, Lin, Liu, Liu, Lu, Lu, Ma, Men, Ren, Ren, Tan, Tan, Tu, Wang, Wang, Wang, Wu, Xu, Xu, Yang, Yang, Yang, Yang, Yao, Yu, Yuan, Yuan, Zhang, Zhang, Zhang, Zhang, Zhou, Zhou, Zhou, and Zhu}]{Qwen2023ABS230916609}
Jinze Bai, Shuai Bai, Yunfei Chu, Zeyu Cui, Kai Dang, Xiaodong Deng, Yang Fan, Wenbin Ge, Yu~Han, Fei Huang, Binyuan Hui, Luo Ji, Mei Li, Junyang Lin, Runji Lin, Dayiheng Liu, Gao Liu, Chengqiang Lu, Keming Lu, Jianxin Ma, Rui Men, Xingzhang Ren, Xuancheng Ren, Chuanqi Tan, Sinan Tan, Jianhong Tu, Peng Wang, Shijie Wang, Wei Wang, Shengguang Wu, Benfeng Xu, Jin Xu, An~Yang, Hao Yang, Jian Yang, Shusheng Yang, Yang Yao, Bowen Yu, Hongyi Yuan, Zheng Yuan, Jianwei Zhang, Xingxuan Zhang, Yichang Zhang, Zhenru Zhang, Chang Zhou, Jingren Zhou, Xiaohuan Zhou, and Tianhang Zhu. 2023.
\newblock \href {https://doi.org/10.48550/ARXIV.2309.16609} {Qwen technical report}.
\newblock \emph{CoRR}, abs/2309.16609.

\bibitem[{Bavarian et~al.(2022)Bavarian, Jun, Tezak, Schulman, McLeavey, Tworek, and Chen}]{FIM2022ABS220714255}
Mohammad Bavarian, Heewoo Jun, Nikolas Tezak, John Schulman, Christine McLeavey, Jerry Tworek, and Mark Chen. 2022.
\newblock \href {https://doi.org/10.48550/ARXIV.2207.14255} {Efficient training of language models to fill in the middle}.
\newblock \emph{CoRR}, abs/2207.14255.

\bibitem[{Bengio et~al.(2009)Bengio, Louradour, Collobert, and Weston}]{CurriculumLearning2009BengioLCW09}
Yoshua Bengio, J{\'{e}}r{\^{o}}me Louradour, Ronan Collobert, and Jason Weston. 2009.
\newblock \href {https://doi.org/10.1145/1553374.1553380} {Curriculum learning}.
\newblock In \emph{Proceedings of the 26th Annual International Conference on Machine Learning, {ICML} 2009, Montreal, Quebec, Canada, June 14-18, 2009}, volume 382 of \emph{{ACM} International Conference Proceeding Series}, pages 41--48. {ACM}.

\bibitem[{Cassano et~al.(2022)Cassano, Gouwar, Nguyen, Nguyen, Phipps{-}Costin, Pinckney, Yee, Zi, Anderson, Feldman, Guha, Greenberg, and Jangda}]{MultiPLE2022ABS220808227}
Federico Cassano, John Gouwar, Daniel Nguyen, Sydney Nguyen, Luna Phipps{-}Costin, Donald Pinckney, Ming{-}Ho Yee, Yangtian Zi, Carolyn~Jane Anderson, Molly~Q. Feldman, Arjun Guha, Michael Greenberg, and Abhinav Jangda. 2022.
\newblock \href {https://doi.org/10.48550/ARXIV.2208.08227} {A scalable and extensible approach to benchmarking nl2code for 18 programming languages}.
\newblock \emph{CoRR}, abs/2208.08227.

\bibitem[{Chaudhary(2023)}]{CodeAlpaca2023GitHub}
Sahil Chaudhary. 2023.
\newblock Code alpaca: An instruction-following llama model for code generation.
\newblock \url{https://github.com/sahil280114/codealpaca}.

\bibitem[{Chen et~al.(2021)Chen, Tworek, Jun, Yuan, de~Oliveira~Pinto, Kaplan, Edwards, Burda, Joseph, Brockman, Ray, Puri, Krueger, Petrov, Khlaaf, Sastry, Mishkin, Chan, Gray, Ryder, Pavlov, Power, Kaiser, Bavarian, Winter, Tillet, Such, Cummings, Plappert, Chantzis, Barnes, Herbert{-}Voss, Guss, Nichol, Paino, Tezak, Tang, Babuschkin, Balaji, Jain, Saunders, Hesse, Carr, Leike, Achiam, Misra, Morikawa, Radford, Knight, Brundage, Murati, Mayer, Welinder, McGrew, Amodei, McCandlish, Sutskever, and Zaremba}]{CodexABS210703374}
Mark Chen, Jerry Tworek, Heewoo Jun, Qiming Yuan, Henrique~Pond{\'{e}} de~Oliveira~Pinto, Jared Kaplan, Harrison Edwards, Yuri Burda, Nicholas Joseph, Greg Brockman, Alex Ray, Raul Puri, Gretchen Krueger, Michael Petrov, Heidy Khlaaf, Girish Sastry, Pamela Mishkin, Brooke Chan, Scott Gray, Nick Ryder, Mikhail Pavlov, Alethea Power, Lukasz Kaiser, Mohammad Bavarian, Clemens Winter, Philippe Tillet, Felipe~Petroski Such, Dave Cummings, Matthias Plappert, Fotios Chantzis, Elizabeth Barnes, Ariel Herbert{-}Voss, William~Hebgen Guss, Alex Nichol, Alex Paino, Nikolas Tezak, Jie Tang, Igor Babuschkin, Suchir Balaji, Shantanu Jain, William Saunders, Christopher Hesse, Andrew~N. Carr, Jan Leike, Joshua Achiam, Vedant Misra, Evan Morikawa, Alec Radford, Matthew Knight, Miles Brundage, Mira Murati, Katie Mayer, Peter Welinder, Bob McGrew, Dario Amodei, Sam McCandlish, Ilya Sutskever, and Wojciech Zaremba. 2021.
\newblock \href {http://arxiv.org/abs/2107.03374} {Evaluating large language models trained on code}.
\newblock \emph{CoRR}, abs/2107.03374.

\bibitem[{Chen et~al.(2022)Chen, Ma, Wang, and Cohen}]{PoT2022ABS221112588}
Wenhu Chen, Xueguang Ma, Xinyi Wang, and William~W. Cohen. 2022.
\newblock \href {https://doi.org/10.48550/ARXIV.2211.12588} {Program of thoughts prompting: Disentangling computation from reasoning for numerical reasoning tasks}.
\newblock \emph{CoRR}, abs/2211.12588.

\bibitem[{Chen et~al.(2023)Chen, Lin, Sch{\"{a}}rli, and Zhou}]{SelfDebug2023ABS230405128}
Xinyun Chen, Maxwell Lin, Nathanael Sch{\"{a}}rli, and Denny Zhou. 2023.
\newblock \href {https://doi.org/10.48550/ARXIV.2304.05128} {Teaching large language models to self-debug}.
\newblock \emph{CoRR}, abs/2304.05128.

\bibitem[{Chowdhery et~al.(2023)Chowdhery, Narang, Devlin, Bosma, Mishra, Roberts, Barham, Chung, Sutton, Gehrmann, Schuh, Shi, Tsvyashchenko, Maynez, Rao, Barnes, Tay, Shazeer, Prabhakaran, Reif, Du, Hutchinson, Pope, Bradbury, Austin, Isard, Gur{-}Ari, Yin, Duke, Levskaya, Ghemawat, Dev, Michalewski, Garcia, Misra, Robinson, Fedus, Zhou, Ippolito, Luan, Lim, Zoph, Spiridonov, Sepassi, Dohan, Agrawal, Omernick, Dai, Pillai, Pellat, Lewkowycz, Moreira, Child, Polozov, Lee, Zhou, Wang, Saeta, Diaz, Firat, Catasta, Wei, Meier{-}Hellstern, Eck, Dean, Petrov, and Fiedel}]{PaLM2023ChowdheryNDBMRBCSGSSTMRBTSPRDHPBAI23}
Aakanksha Chowdhery, Sharan Narang, Jacob Devlin, Maarten Bosma, Gaurav Mishra, Adam Roberts, Paul Barham, Hyung~Won Chung, Charles Sutton, Sebastian Gehrmann, Parker Schuh, Kensen Shi, Sasha Tsvyashchenko, Joshua Maynez, Abhishek Rao, Parker Barnes, Yi~Tay, Noam Shazeer, Vinodkumar Prabhakaran, Emily Reif, Nan Du, Ben Hutchinson, Reiner Pope, James Bradbury, Jacob Austin, Michael Isard, Guy Gur{-}Ari, Pengcheng Yin, Toju Duke, Anselm Levskaya, Sanjay Ghemawat, Sunipa Dev, Henryk Michalewski, Xavier Garcia, Vedant Misra, Kevin Robinson, Liam Fedus, Denny Zhou, Daphne Ippolito, David Luan, Hyeontaek Lim, Barret Zoph, Alexander Spiridonov, Ryan Sepassi, David Dohan, Shivani Agrawal, Mark Omernick, Andrew~M. Dai, Thanumalayan~Sankaranarayana Pillai, Marie Pellat, Aitor Lewkowycz, Erica Moreira, Rewon Child, Oleksandr Polozov, Katherine Lee, Zongwei Zhou, Xuezhi Wang, Brennan Saeta, Mark Diaz, Orhan Firat, Michele Catasta, Jason Wei, Kathy Meier{-}Hellstern, Douglas Eck, Jeff Dean, Slav Petrov, and Noah Fiedel.
  2023.
\newblock \href {http://jmlr.org/papers/v24/22-1144.html} {Palm: Scaling language modeling with pathways}.
\newblock \emph{J. Mach. Learn. Res.}, 24:240:1--240:113.

\bibitem[{Cobbe et~al.(2021)Cobbe, Kosaraju, Bavarian, Chen, Jun, Kaiser, Plappert, Tworek, Hilton, Nakano, Hesse, and Schulman}]{GSM8k2021Karl}
Karl Cobbe, Vineet Kosaraju, Mohammad Bavarian, Mark Chen, Heewoo Jun, Lukasz Kaiser, Matthias Plappert, Jerry Tworek, Jacob Hilton, Reiichiro Nakano, Christopher Hesse, and John Schulman. 2021.
\newblock \href {http://arxiv.org/abs/2110.14168} {Training verifiers to solve math word problems}.
\newblock \emph{CoRR}, abs/2110.14168.

\bibitem[{Dao(2023)}]{FlashAttention2023ABS230708691}
Tri Dao. 2023.
\newblock \href {https://doi.org/10.48550/ARXIV.2307.08691} {Flashattention-2: Faster attention with better parallelism and work partitioning}.
\newblock \emph{CoRR}, abs/2307.08691.

\bibitem[{Devlin et~al.(2019)Devlin, Chang, Lee, and Toutanova}]{BERT2019DevlinCLT19}
Jacob Devlin, Ming{-}Wei Chang, Kenton Lee, and Kristina Toutanova. 2019.
\newblock \href {https://doi.org/10.18653/V1/N19-1423} {{BERT:} pre-training of deep bidirectional transformers for language understanding}.
\newblock In \emph{Proceedings of the 2019 Conference of the North American Chapter of the Association for Computational Linguistics: Human Language Technologies, {NAACL-HLT} 2019, Minneapolis, MN, USA, June 2-7, 2019, Volume 1 (Long and Short Papers)}, pages 4171--4186. Association for Computational Linguistics.

\bibitem[{Dou et~al.(2024)Dou, Liu, Jia, Xiong, Zhou, Shen, Shan, Huang, Wang, Fan, Xi, Zhou, Ji, Zheng, Zhang, Huang, and Gui}]{StepCoder2024ABS240201391}
Shihan Dou, Yan Liu, Haoxiang Jia, Limao Xiong, Enyu Zhou, Wei Shen, Junjie Shan, Caishuang Huang, Xiao Wang, Xiaoran Fan, Zhiheng Xi, Yuhao Zhou, Tao Ji, Rui Zheng, Qi~Zhang, Xuanjing Huang, and Tao Gui. 2024.
\newblock \href {https://doi.org/10.48550/ARXIV.2402.01391} {Stepcoder: Improve code generation with reinforcement learning from compiler feedback}.
\newblock \emph{CoRR}, abs/2402.01391.

\bibitem[{Du et~al.(2023)Du, Liu, Wang, Wang, Liu, Chen, Feng, Sha, Peng, and Lou}]{ClassEval2023Xueying}
Xueying Du, Mingwei Liu, Kaixin Wang, Hanlin Wang, Junwei Liu, Yixuan Chen, Jiayi Feng, Chaofeng Sha, Xin Peng, and Yiling Lou. 2023.
\newblock \href {https://doi.org/10.48550/ARXIV.2308.01861} {Classeval: {A} manually-crafted benchmark for evaluating llms on class-level code generation}.
\newblock \emph{CoRR}, abs/2308.01861.

\bibitem[{Dubey et~al.(2024)Dubey, Jauhri, Pandey, Kadian, Al-Dahle, Letman, Mathur, Schelten, Yang, Fan et~al.}]{Llama312024Dubey}
Abhimanyu Dubey, Abhinav Jauhri, Abhinav Pandey, Abhishek Kadian, Ahmad Al-Dahle, Aiesha Letman, Akhil Mathur, Alan Schelten, Amy Yang, Angela Fan, et~al. 2024.
\newblock The llama 3 herd of models.
\newblock \emph{arXiv preprint arXiv:2407.21783}.

\bibitem[{Feng et~al.(2020)Feng, Guo, Tang, Duan, Feng, Gong, Shou, Qin, Liu, Jiang, and Zhou}]{CodeBertFengGTDFGS0LJZ20}
Zhangyin Feng, Daya Guo, Duyu Tang, Nan Duan, Xiaocheng Feng, Ming Gong, Linjun Shou, Bing Qin, Ting Liu, Daxin Jiang, and Ming Zhou. 2020.
\newblock \href {https://doi.org/10.18653/V1/2020.FINDINGS-EMNLP.139} {Codebert: {A} pre-trained model for programming and natural languages}.
\newblock In \emph{Findings of the Association for Computational Linguistics: {EMNLP} 2020, Online Event, 16-20 November 2020}, volume {EMNLP} 2020 of \emph{Findings of {ACL}}, pages 1536--1547. Association for Computational Linguistics.

\bibitem[{Fried et~al.(2023)Fried, Aghajanyan, Lin, Wang, Wallace, Shi, Zhong, Yih, Zettlemoyer, and Lewis}]{InCoderFriedAL0WSZYZL23}
Daniel Fried, Armen Aghajanyan, Jessy Lin, Sida Wang, Eric Wallace, Freda Shi, Ruiqi Zhong, Scott Yih, Luke Zettlemoyer, and Mike Lewis. 2023.
\newblock \href {https://openreview.net/pdf?id=hQwb-lbM6EL} {Incoder: {A} generative model for code infilling and synthesis}.
\newblock In \emph{The Eleventh International Conference on Learning Representations, {ICLR} 2023, Kigali, Rwanda, May 1-5, 2023}. OpenReview.net.

\bibitem[{Gao et~al.(2023)Gao, Madaan, Zhou, Alon, Liu, Yang, Callan, and Neubig}]{PAL2023GaoMZ00YCN23}
Luyu Gao, Aman Madaan, Shuyan Zhou, Uri Alon, Pengfei Liu, Yiming Yang, Jamie Callan, and Graham Neubig. 2023.
\newblock \href {https://proceedings.mlr.press/v202/gao23f.html} {{PAL:} program-aided language models}.
\newblock In \emph{International Conference on Machine Learning, {ICML} 2023, 23-29 July 2023, Honolulu, Hawaii, {USA}}, volume 202 of \emph{Proceedings of Machine Learning Research}, pages 10764--10799. {PMLR}.

\bibitem[{Gong et~al.(2024)Gong, Elhoushi, and Cheung}]{ASTT52024ABS240103003}
Linyuan Gong, Mostafa Elhoushi, and Alvin Cheung. 2024.
\newblock \href {https://doi.org/10.48550/ARXIV.2401.03003} {{AST-T5:} structure-aware pretraining for code generation and understanding}.
\newblock \emph{CoRR}, abs/2401.03003.

\bibitem[{Guo et~al.(2024)Guo, Zhu, Yang, Xie, Dong, Zhang, Chen, Bi, Wu, Li, Luo, Xiong, and Liang}]{DeepSeekCoderABS240114196}
Daya Guo, Qihao Zhu, Dejian Yang, Zhenda Xie, Kai Dong, Wentao Zhang, Guanting Chen, Xiao Bi, Y.~Wu, Y.~K. Li, Fuli Luo, Yingfei Xiong, and Wenfeng Liang. 2024.
\newblock \href {https://doi.org/10.48550/ARXIV.2401.14196} {Deepseek-coder: When the large language model meets programming - the rise of code intelligence}.
\newblock \emph{CoRR}, abs/2401.14196.

\bibitem[{Hendrycks et~al.(2021{\natexlab{a}})Hendrycks, Basart, Kadavath, Mazeika, Arora, Guo, Burns, Puranik, He, Song, and Steinhardt}]{APPs2021Dan}
Dan Hendrycks, Steven Basart, Saurav Kadavath, Mantas Mazeika, Akul Arora, Ethan Guo, Collin Burns, Samir Puranik, Horace He, Dawn Song, and Jacob Steinhardt. 2021{\natexlab{a}}.
\newblock \href {https://datasets-benchmarks-proceedings.neurips.cc/paper/2021/hash/c24cd76e1ce41366a4bbe8a49b02a028-Abstract-round2.html} {Measuring coding challenge competence with {APPS}}.
\newblock In \emph{Proceedings of the Neural Information Processing Systems Track on Datasets and Benchmarks 1, NeurIPS Datasets and Benchmarks 2021, December 2021, virtual}.

\bibitem[{Hendrycks et~al.(2021{\natexlab{b}})Hendrycks, Burns, Kadavath, Arora, Basart, Tang, Song, and Steinhardt}]{Math2021Dan}
Dan Hendrycks, Collin Burns, Saurav Kadavath, Akul Arora, Steven Basart, Eric Tang, Dawn Song, and Jacob Steinhardt. 2021{\natexlab{b}}.
\newblock \href {https://datasets-benchmarks-proceedings.neurips.cc/paper/2021/hash/be83ab3ecd0db773eb2dc1b0a17836a1-Abstract-round2.html} {Measuring mathematical problem solving with the {MATH} dataset}.
\newblock In \emph{Proceedings of the Neural Information Processing Systems Track on Datasets and Benchmarks 1, NeurIPS Datasets and Benchmarks 2021, December 2021, virtual}.

\bibitem[{Hu et~al.(2024)Hu, Kuang, Sun, Yang, and Wu}]{PrintDebugABS240105319}
Xueyu Hu, Kun Kuang, Jiankai Sun, Hongxia Yang, and Fei Wu. 2024.
\newblock \href {https://doi.org/10.48550/ARXIV.2401.05319} {Leveraging print debugging to improve code generation in large language models}.
\newblock \emph{CoRR}, abs/2401.05319.

\bibitem[{Jaech et~al.(2024)Jaech, Kalai, Lerer, Richardson, El{-}Kishky, Low, Helyar, Madry, Beutel, Carney, Iftimie, Karpenko, Passos, Neitz, Prokofiev, Wei, Tam, Bennett, Kumar, Saraiva, Vallone, Duberstein, Kondrich, Mishchenko, Applebaum, Jiang, Nair, Zoph, Ghorbani, Rossen, Sokolowsky, Barak, McGrew, Minaiev, Hao, Baker, Houghton, McKinzie, Eastman, Lugaresi, Bassin, Hudson, Li, de~Bourcy, Voss, Shen, Zhang, Koch, Orsinger, Hesse, Fischer, Chan, Roberts, Kappler, Levy, Selsam, Dohan, Farhi, Mely, Robinson, Tsipras, Li, Oprica, Freeman, Zhang, Wong, Proehl, Cheung, Mitchell, Wallace, Ritter, Mays, Wang, Such, Raso, Leoni, Tsimpourlas, Song, von Lohmann, Sulit, Salmon, Parascandolo, Chabot, Zhao, Brockman, Leclerc, Salman, Bao, Sheng, Andrin, Bagherinezhad, Ren, Lightman, Chung, Kivlichan, O'Connell, Osband, Gilaberte, and Akkaya}]{O12024Aaron}
Aaron Jaech, Adam Kalai, Adam Lerer, Adam Richardson, Ahmed El{-}Kishky, Aiden Low, Alec Helyar, Aleksander Madry, Alex Beutel, Alex Carney, Alex Iftimie, Alex Karpenko, Alex~Tachard Passos, Alexander Neitz, Alexander Prokofiev, Alexander Wei, Allison Tam, Ally Bennett, Ananya Kumar, Andre Saraiva, Andrea Vallone, Andrew Duberstein, Andrew Kondrich, Andrey Mishchenko, Andy Applebaum, Angela Jiang, Ashvin Nair, Barret Zoph, Behrooz Ghorbani, Ben Rossen, Benjamin Sokolowsky, Boaz Barak, Bob McGrew, Borys Minaiev, Botao Hao, Bowen Baker, Brandon Houghton, Brandon McKinzie, Brydon Eastman, Camillo Lugaresi, Cary Bassin, Cary Hudson, Chak~Ming Li, Charles de~Bourcy, Chelsea Voss, Chen Shen, Chong Zhang, Chris Koch, Chris Orsinger, Christopher Hesse, Claudia Fischer, Clive Chan, Dan Roberts, Daniel Kappler, Daniel Levy, Daniel Selsam, David Dohan, David Farhi, David Mely, David Robinson, Dimitris Tsipras, Doug Li, Dragos Oprica, Eben Freeman, Eddie Zhang, Edmund Wong, Elizabeth Proehl, Enoch Cheung, Eric Mitchell,
  Eric Wallace, Erik Ritter, Evan Mays, Fan Wang, Felipe~Petroski Such, Filippo Raso, Florencia Leoni, Foivos Tsimpourlas, Francis Song, Fred von Lohmann, Freddie Sulit, Geoff Salmon, Giambattista Parascandolo, Gildas Chabot, Grace Zhao, Greg Brockman, Guillaume Leclerc, Hadi Salman, Haiming Bao, Hao Sheng, Hart Andrin, Hessam Bagherinezhad, Hongyu Ren, Hunter Lightman, Hyung~Won Chung, Ian Kivlichan, Ian O'Connell, Ian Osband, Ignasi~Clavera Gilaberte, and Ilge Akkaya. 2024.
\newblock \href {https://doi.org/10.48550/ARXIV.2412.16720} {Openai o1 system card}.
\newblock \emph{CoRR}, abs/2412.16720.

\bibitem[{Jain et~al.(2024)Jain, Han, Gu, Li, Yan, Zhang, Wang, Solar{-}Lezama, Sen, and Stoica}]{LiveCodeBench2024Naman}
Naman Jain, King Han, Alex Gu, Wen{-}Ding Li, Fanjia Yan, Tianjun Zhang, Sida Wang, Armando Solar{-}Lezama, Koushik Sen, and Ion Stoica. 2024.
\newblock \href {https://doi.org/10.48550/ARXIV.2403.07974} {Livecodebench: Holistic and contamination free evaluation of large language models for code}.
\newblock \emph{CoRR}, abs/2403.07974.

\bibitem[{Jiang et~al.(2023)Jiang, Sablayrolles, Mensch, Bamford, Chaplot, de~Las~Casas, Bressand, Lengyel, Lample, Saulnier, Lavaud, Lachaux, Stock, Scao, Lavril, Wang, Lacroix, and Sayed}]{Mistral2023ABS231006825}
Albert~Q. Jiang, Alexandre Sablayrolles, Arthur Mensch, Chris Bamford, Devendra~Singh Chaplot, Diego de~Las~Casas, Florian Bressand, Gianna Lengyel, Guillaume Lample, Lucile Saulnier, L{\'{e}}lio~Renard Lavaud, Marie{-}Anne Lachaux, Pierre Stock, Teven~Le Scao, Thibaut Lavril, Thomas Wang, Timoth{\'{e}}e Lacroix, and William~El Sayed. 2023.
\newblock \href {https://doi.org/10.48550/ARXIV.2310.06825} {Mistral 7b}.
\newblock \emph{CoRR}, abs/2310.06825.

\bibitem[{Jiang et~al.(2024)Jiang, Sablayrolles, Roux, Mensch, Savary, Bamford, Chaplot, de~Las~Casas, Hanna, Bressand, Lengyel, Bour, Lample, Lavaud, Saulnier, Lachaux, Stock, Subramanian, Yang, Antoniak, Scao, Gervet, Lavril, Wang, Lacroix, and Sayed}]{Mixtral2023ABS240104088}
Albert~Q. Jiang, Alexandre Sablayrolles, Antoine Roux, Arthur Mensch, Blanche Savary, Chris Bamford, Devendra~Singh Chaplot, Diego de~Las~Casas, Emma~Bou Hanna, Florian Bressand, Gianna Lengyel, Guillaume Bour, Guillaume Lample, L{\'{e}}lio~Renard Lavaud, Lucile Saulnier, Marie{-}Anne Lachaux, Pierre Stock, Sandeep Subramanian, Sophia Yang, Szymon Antoniak, Teven~Le Scao, Th{\'{e}}ophile Gervet, Thibaut Lavril, Thomas Wang, Timoth{\'{e}}e Lacroix, and William~El Sayed. 2024.
\newblock \href {https://doi.org/10.48550/ARXIV.2401.04088} {Mixtral of experts}.
\newblock \emph{CoRR}, abs/2401.04088.

\bibitem[{Lai et~al.(2023)Lai, Li, Wang, Zhang, Zhong, Zettlemoyer, Yih, Fried, Wang, and Yu}]{DS10002023Lai0WZZZYFWY23}
Yuhang Lai, Chengxi Li, Yiming Wang, Tianyi Zhang, Ruiqi Zhong, Luke Zettlemoyer, Wen{-}Tau Yih, Daniel Fried, Sida~I. Wang, and Tao Yu. 2023.
\newblock \href {https://proceedings.mlr.press/v202/lai23b.html} {{DS-1000:} {A} natural and reliable benchmark for data science code generation}.
\newblock In \emph{International Conference on Machine Learning, {ICML} 2023, 23-29 July 2023, Honolulu, Hawaii, {USA}}, volume 202 of \emph{Proceedings of Machine Learning Research}, pages 18319--18345. {PMLR}.

\bibitem[{Lei et~al.(2024)Lei, Li, and Chen}]{AutoCoder2024Bin}
Bin Lei, Yuchen Li, and Qiuwu Chen. 2024.
\newblock \href {https://doi.org/10.48550/ARXIV.2405.14906} {Autocoder: Enhancing code large language model with aiev-instruct}.
\newblock \emph{CoRR}, abs/2405.14906.

\bibitem[{Li et~al.(2023)Li, Allal, Zi, Muennighoff, Kocetkov, Mou, Marone, Akiki, Li, Chim, Liu, Zheltonozhskii, Zhuo, Wang, Dehaene, Davaadorj, Lamy{-}Poirier, Monteiro, Shliazhko, Gontier, Meade, Zebaze, Yee, Umapathi, Zhu, Lipkin, Oblokulov, Wang, V, Stillerman, Patel, Abulkhanov, Zocca, Dey, Zhang, Moustafa{-}Fahmy, Bhattacharyya, Yu, Singh, Luccioni, Villegas, Kunakov, Zhdanov, Romero, Lee, Timor, Ding, Schlesinger, Schoelkopf, Ebert, Dao, Mishra, Gu, Robinson, Anderson, Dolan{-}Gavitt, Contractor, Reddy, Fried, Bahdanau, Jernite, Ferrandis, Hughes, Wolf, Guha, von Werra, and de~Vries}]{StarCoderABS230506161}
Raymond Li, Loubna~Ben Allal, Yangtian Zi, Niklas Muennighoff, Denis Kocetkov, Chenghao Mou, Marc Marone, Christopher Akiki, Jia Li, Jenny Chim, Qian Liu, Evgenii Zheltonozhskii, Terry~Yue Zhuo, Thomas Wang, Olivier Dehaene, Mishig Davaadorj, Joel Lamy{-}Poirier, Jo{\~{a}}o Monteiro, Oleh Shliazhko, Nicolas Gontier, Nicholas Meade, Armel Zebaze, Ming{-}Ho Yee, Logesh~Kumar Umapathi, Jian Zhu, Benjamin Lipkin, Muhtasham Oblokulov, Zhiruo Wang, Rudra~Murthy V, Jason Stillerman, Siva~Sankalp Patel, Dmitry Abulkhanov, Marco Zocca, Manan Dey, Zhihan Zhang, Nour Moustafa{-}Fahmy, Urvashi Bhattacharyya, Wenhao Yu, Swayam Singh, Sasha Luccioni, Paulo Villegas, Maxim Kunakov, Fedor Zhdanov, Manuel Romero, Tony Lee, Nadav Timor, Jennifer Ding, Claire Schlesinger, Hailey Schoelkopf, Jan Ebert, Tri Dao, Mayank Mishra, Alex Gu, Jennifer Robinson, Carolyn~Jane Anderson, Brendan Dolan{-}Gavitt, Danish Contractor, Siva Reddy, Daniel Fried, Dzmitry Bahdanau, Yacine Jernite, Carlos~Mu{\~{n}}oz Ferrandis, Sean Hughes, Thomas
  Wolf, Arjun Guha, Leandro von Werra, and Harm de~Vries. 2023.
\newblock \href {https://doi.org/10.48550/ARXIV.2305.06161} {Starcoder: may the source be with you!}
\newblock \emph{CoRR}, abs/2305.06161.

\bibitem[{Liu et~al.(2023)Liu, Xia, Wang, and Zhang}]{EvalPlus2023LiuXW023}
Jiawei Liu, Chunqiu~Steven Xia, Yuyao Wang, and Lingming Zhang. 2023.
\newblock \href {http://papers.nips.cc/paper\_files/paper/2023/hash/43e9d647ccd3e4b7b5baab53f0368686-Abstract-Conference.html} {Is your code generated by chatgpt really correct? rigorous evaluation of large language models for code generation}.
\newblock In \emph{Advances in Neural Information Processing Systems 36: Annual Conference on Neural Information Processing Systems 2023, NeurIPS 2023, New Orleans, LA, USA, December 10 - 16, 2023}.

\bibitem[{Longpre et~al.(2023)Longpre, Hou, Vu, Webson, Chung, Tay, Zhou, Le, Zoph, Wei, and Roberts}]{Flan2023LongpreHVWCTZLZ23}
Shayne Longpre, Le~Hou, Tu~Vu, Albert Webson, Hyung~Won Chung, Yi~Tay, Denny Zhou, Quoc~V. Le, Barret Zoph, Jason Wei, and Adam Roberts. 2023.
\newblock \href {https://proceedings.mlr.press/v202/longpre23a.html} {The flan collection: Designing data and methods for effective instruction tuning}.
\newblock In \emph{International Conference on Machine Learning, {ICML} 2023, 23-29 July 2023, Honolulu, Hawaii, {USA}}, volume 202 of \emph{Proceedings of Machine Learning Research}, pages 22631--22648. {PMLR}.

\bibitem[{Loshchilov and Hutter(2019)}]{AdamW2019LoshchilovH19}
Ilya Loshchilov and Frank Hutter. 2019.
\newblock \href {https://openreview.net/forum?id=Bkg6RiCqY7} {Decoupled weight decay regularization}.
\newblock In \emph{7th International Conference on Learning Representations, {ICLR} 2019, New Orleans, LA, USA, May 6-9, 2019}. OpenReview.net.

\bibitem[{Lozhkov et~al.(2024)Lozhkov, Li, Allal, Cassano, Lamy-Poirier, Tazi, Tang, Pykhtar, Liu, Wei, Liu, Tian, Kocetkov, Zucker, Belkada, Wang, Liu, Abulkhanov, Paul, Li, Li, Risdal, Li, Zhu, Zhuo, Zheltonozhskii, Dade, Yu, Krauß, Jain, Su, He, Dey, Abati, Chai, Muennighoff, Tang, Oblokulov, Akiki, Marone, Mou, Mishra, Gu, Hui, Dao, Zebaze, Dehaene, Patry, Xu, McAuley, Hu, Scholak, Paquet, Robinson, Anderson, Chapados, Patwary, Tajbakhsh, Jernite, Ferrandis, Zhang, Hughes, Wolf, Guha, von Werra, and de~Vries}]{StarCoder22024ABS240219173}
Anton Lozhkov, Raymond Li, Loubna~Ben Allal, Federico Cassano, Joel Lamy-Poirier, Nouamane Tazi, Ao~Tang, Dmytro Pykhtar, Jiawei Liu, Yuxiang Wei, Tianyang Liu, Max Tian, Denis Kocetkov, Arthur Zucker, Younes Belkada, Zijian Wang, Qian Liu, Dmitry Abulkhanov, Indraneil Paul, Zhuang Li, Wen-Ding Li, Megan Risdal, Jia Li, Jian Zhu, Terry~Yue Zhuo, Evgenii Zheltonozhskii, Nii Osae~Osae Dade, Wenhao Yu, Lucas Krauß, Naman Jain, Yixuan Su, Xuanli He, Manan Dey, Edoardo Abati, Yekun Chai, Niklas Muennighoff, Xiangru Tang, Muhtasham Oblokulov, Christopher Akiki, Marc Marone, Chenghao Mou, Mayank Mishra, Alex Gu, Binyuan Hui, Tri Dao, Armel Zebaze, Olivier Dehaene, Nicolas Patry, Canwen Xu, Julian McAuley, Han Hu, Torsten Scholak, Sebastien Paquet, Jennifer Robinson, Carolyn~Jane Anderson, Nicolas Chapados, Mostofa Patwary, Nima Tajbakhsh, Yacine Jernite, Carlos~Muñoz Ferrandis, Lingming Zhang, Sean Hughes, Thomas Wolf, Arjun Guha, Leandro von Werra, and Harm de~Vries. 2024.
\newblock \href {https://doi.org/10.48550/ARXIV.2402.19173} {Starcoder 2 and the stack v2: The next generation}.
\newblock \emph{CoRR}, abs/2402.19173.

\bibitem[{Luo et~al.(2023)Luo, Xu, Zhao, Sun, Geng, Hu, Tao, Ma, Lin, and Jiang}]{WizardCoder2023ABS230608568}
Ziyang Luo, Can Xu, Pu~Zhao, Qingfeng Sun, Xiubo Geng, Wenxiang Hu, Chongyang Tao, Jing Ma, Qingwei Lin, and Daxin Jiang. 2023.
\newblock \href {https://doi.org/10.48550/ARXIV.2306.08568} {Wizardcoder: Empowering code large language models with evol-instruct}.
\newblock \emph{CoRR}, abs/2306.08568.

\bibitem[{Madaan et~al.(2023)Madaan, Tandon, Gupta, Hallinan, Gao, Wiegreffe, Alon, Dziri, Prabhumoye, Yang, Gupta, Majumder, Hermann, Welleck, Yazdanbakhsh, and Clark}]{SelfRefine2023MadaanTGHGW0DPY23}
Aman Madaan, Niket Tandon, Prakhar Gupta, Skyler Hallinan, Luyu Gao, Sarah Wiegreffe, Uri Alon, Nouha Dziri, Shrimai Prabhumoye, Yiming Yang, Shashank Gupta, Bodhisattwa~Prasad Majumder, Katherine Hermann, Sean Welleck, Amir Yazdanbakhsh, and Peter Clark. 2023.
\newblock \href {http://papers.nips.cc/paper\_files/paper/2023/hash/91edff07232fb1b55a505a9e9f6c0ff3-Abstract-Conference.html} {Self-refine: Iterative refinement with self-feedback}.
\newblock In \emph{Advances in Neural Information Processing Systems 36: Annual Conference on Neural Information Processing Systems 2023, NeurIPS 2023, New Orleans, LA, USA, December 10 - 16, 2023}.

\bibitem[{Mirzadeh et~al.(2020)Mirzadeh, Farajtabar, Li, Levine, Matsukawa, and Ghasemzadeh}]{Assistant2020MirzadehFLLMG20}
Seyed{-}Iman Mirzadeh, Mehrdad Farajtabar, Ang Li, Nir Levine, Akihiro Matsukawa, and Hassan Ghasemzadeh. 2020.
\newblock \href {https://doi.org/10.1609/AAAI.V34I04.5963} {Improved knowledge distillation via teacher assistant}.
\newblock In \emph{The Thirty-Fourth {AAAI} Conference on Artificial Intelligence, {AAAI} 2020, The Thirty-Second Innovative Applications of Artificial Intelligence Conference, {IAAI} 2020, The Tenth {AAAI} Symposium on Educational Advances in Artificial Intelligence, {EAAI} 2020, New York, NY, USA, February 7-12, 2020}, pages 5191--5198. {AAAI} Press.

\bibitem[{Muennighoff et~al.(2023)Muennighoff, Liu, Zebaze, Zheng, Hui, Zhuo, Singh, Tang, von Werra, and Longpre}]{OctoPack2023ABS230807124}
Niklas Muennighoff, Qian Liu, Armel Zebaze, Qinkai Zheng, Binyuan Hui, Terry~Yue Zhuo, Swayam Singh, Xiangru Tang, Leandro von Werra, and Shayne Longpre. 2023.
\newblock \href {https://doi.org/10.48550/ARXIV.2308.07124} {Octopack: Instruction tuning code large language models}.
\newblock \emph{CoRR}, abs/2308.07124.

\bibitem[{Nijkamp et~al.(2023)Nijkamp, Pang, Hayashi, Tu, Wang, Zhou, Savarese, and Xiong}]{CodeGen2023NijkampPHTWZSX23}
Erik Nijkamp, Bo~Pang, Hiroaki Hayashi, Lifu Tu, Huan Wang, Yingbo Zhou, Silvio Savarese, and Caiming Xiong. 2023.
\newblock \href {https://openreview.net/pdf?id=iaYcJKpY2B\_} {Codegen: An open large language model for code with multi-turn program synthesis}.
\newblock In \emph{The Eleventh International Conference on Learning Representations, {ICLR} 2023, Kigali, Rwanda, May 1-5, 2023}. OpenReview.net.

\bibitem[{OpenAI(2023)}]{GPT42023ABS230308774}
OpenAI. 2023.
\newblock \href {https://doi.org/10.48550/ARXIV.2303.08774} {{GPT-4} technical report}.
\newblock \emph{CoRR}, abs/2303.08774.

\bibitem[{Ouyang et~al.(2022)Ouyang, Wu, Jiang, Almeida, Wainwright, Mishkin, Zhang, Agarwal, Slama, Ray, Schulman, Hilton, Kelton, Miller, Simens, Askell, Welinder, Christiano, Leike, and Lowe}]{InstuctGPT2022Ouyang0JAWMZASR22}
Long Ouyang, Jeffrey Wu, Xu~Jiang, Diogo Almeida, Carroll~L. Wainwright, Pamela Mishkin, Chong Zhang, Sandhini Agarwal, Katarina Slama, Alex Ray, John Schulman, Jacob Hilton, Fraser Kelton, Luke Miller, Maddie Simens, Amanda Askell, Peter Welinder, Paul~F. Christiano, Jan Leike, and Ryan Lowe. 2022.
\newblock \href {http://papers.nips.cc/paper\_files/paper/2022/hash/b1efde53be364a73914f58805a001731-Abstract-Conference.html} {Training language models to follow instructions with human feedback}.
\newblock In \emph{Advances in Neural Information Processing Systems 35: Annual Conference on Neural Information Processing Systems 2022, NeurIPS 2022, New Orleans, LA, USA, November 28 - December 9, 2022}.

\bibitem[{Penedo et~al.(2023)Penedo, Malartic, Hesslow, Cojocaru, Alobeidli, Cappelli, Pannier, Almazrouei, and Launay}]{Falcon2023PenedoMHCACPAL23}
Guilherme Penedo, Quentin Malartic, Daniel Hesslow, Ruxandra Cojocaru, Hamza Alobeidli, Alessandro Cappelli, Baptiste Pannier, Ebtesam Almazrouei, and Julien Launay. 2023.
\newblock \href {http://papers.nips.cc/paper\_files/paper/2023/hash/fa3ed726cc5073b9c31e3e49a807789c-Abstract-Datasets\_and\_Benchmarks.html} {The refinedweb dataset for falcon {LLM:} outperforming curated corpora with web data only}.
\newblock In \emph{Advances in Neural Information Processing Systems 36: Annual Conference on Neural Information Processing Systems 2023, NeurIPS 2023, New Orleans, LA, USA, December 10 - 16, 2023}.

\bibitem[{Raffel et~al.(2020)Raffel, Shazeer, Roberts, Lee, Narang, Matena, Zhou, Li, and Liu}]{T52020RaffelSRLNMZLL20}
Colin Raffel, Noam Shazeer, Adam Roberts, Katherine Lee, Sharan Narang, Michael Matena, Yanqi Zhou, Wei Li, and Peter~J. Liu. 2020.
\newblock \href {http://jmlr.org/papers/v21/20-074.html} {Exploring the limits of transfer learning with a unified text-to-text transformer}.
\newblock \emph{J. Mach. Learn. Res.}, 21:140:1--140:67.

\bibitem[{Rajbhandari et~al.(2020)Rajbhandari, Rasley, Ruwase, and He}]{ZERO2020RajbhandariRRH20}
Samyam Rajbhandari, Jeff Rasley, Olatunji Ruwase, and Yuxiong He. 2020.
\newblock \href {https://doi.org/10.1109/SC41405.2020.00024} {Zero: memory optimizations toward training trillion parameter models}.
\newblock In \emph{Proceedings of the International Conference for High Performance Computing, Networking, Storage and Analysis, {SC} 2020, Virtual Event / Atlanta, Georgia, USA, November 9-19, 2020}, page~20. {IEEE/ACM}.

\bibitem[{Rozi{\`{e}}re et~al.(2023)Rozi{\`{e}}re, Gehring, Gloeckle, Sootla, Gat, Tan, Adi, Liu, Remez, Rapin, Kozhevnikov, Evtimov, Bitton, Bhatt, Canton{-}Ferrer, Grattafiori, Xiong, D{\'{e}}fossez, Copet, Azhar, Touvron, Martin, Usunier, Scialom, and Synnaeve}]{CodeLlamaABS230812950}
Baptiste Rozi{\`{e}}re, Jonas Gehring, Fabian Gloeckle, Sten Sootla, Itai Gat, Xiaoqing~Ellen Tan, Yossi Adi, Jingyu Liu, Tal Remez, J{\'{e}}r{\'{e}}my Rapin, Artyom Kozhevnikov, Ivan Evtimov, Joanna Bitton, Manish Bhatt, Cristian Canton{-}Ferrer, Aaron Grattafiori, Wenhan Xiong, Alexandre D{\'{e}}fossez, Jade Copet, Faisal Azhar, Hugo Touvron, Louis Martin, Nicolas Usunier, Thomas Scialom, and Gabriel Synnaeve. 2023.
\newblock \href {https://doi.org/10.48550/ARXIV.2308.12950} {Code llama: Open foundation models for code}.
\newblock \emph{CoRR}, abs/2308.12950.

\bibitem[{Sanh et~al.(2022)Sanh, Webson, Raffel, Bach, Sutawika, Alyafeai, Chaffin, Stiegler, Raja, Dey, Bari, Xu, Thakker, Sharma, Szczechla, Kim, Chhablani, Nayak, Datta, Chang, Jiang, Wang, Manica, Shen, Yong, Pandey, Bawden, Wang, Neeraj, Rozen, Sharma, Santilli, F{\'{e}}vry, Fries, Teehan, Scao, Biderman, Gao, Wolf, and Rush}]{MultiTaskPrompt2022SanhWRBSACSRDBX22}
Victor Sanh, Albert Webson, Colin Raffel, Stephen~H. Bach, Lintang Sutawika, Zaid Alyafeai, Antoine Chaffin, Arnaud Stiegler, Arun Raja, Manan Dey, M~Saiful Bari, Canwen Xu, Urmish Thakker, Shanya~Sharma Sharma, Eliza Szczechla, Taewoon Kim, Gunjan Chhablani, Nihal~V. Nayak, Debajyoti Datta, Jonathan Chang, Mike~Tian{-}Jian Jiang, Han Wang, Matteo Manica, Sheng Shen, Zheng~Xin Yong, Harshit Pandey, Rachel Bawden, Thomas Wang, Trishala Neeraj, Jos Rozen, Abheesht Sharma, Andrea Santilli, Thibault F{\'{e}}vry, Jason~Alan Fries, Ryan Teehan, Teven~Le Scao, Stella Biderman, Leo Gao, Thomas Wolf, and Alexander~M. Rush. 2022.
\newblock \href {https://openreview.net/forum?id=9Vrb9D0WI4} {Multitask prompted training enables zero-shot task generalization}.
\newblock In \emph{The Tenth International Conference on Learning Representations, {ICLR} 2022, Virtual Event, April 25-29, 2022}. OpenReview.net.

\bibitem[{Shen et~al.(2023)Shen, Zhang, Chen, Zan, Geng, Fu, Zeng, Yu, Ji, Zhao, Guo, and Wang}]{PanGuCoder22023ABS230714936}
Bo~Shen, Jiaxin Zhang, Taihong Chen, Daoguang Zan, Bing Geng, An~Fu, Muhan Zeng, Ailun Yu, Jichuan Ji, Jingyang Zhao, Yuenan Guo, and Qianxiang Wang. 2023.
\newblock \href {https://doi.org/10.48550/ARXIV.2307.14936} {Pangu-coder2: Boosting large language models for code with ranking feedback}.
\newblock \emph{CoRR}, abs/2307.14936.

\bibitem[{Shinn et~al.(2023)Shinn, Cassano, Gopinath, Narasimhan, and Yao}]{Reflexion2023ShinnCGNY23}
Noah Shinn, Federico Cassano, Ashwin Gopinath, Karthik Narasimhan, and Shunyu Yao. 2023.
\newblock \href {http://papers.nips.cc/paper\_files/paper/2023/hash/1b44b878bb782e6954cd888628510e90-Abstract-Conference.html} {Reflexion: language agents with verbal reinforcement learning}.
\newblock In \emph{Advances in Neural Information Processing Systems 36: Annual Conference on Neural Information Processing Systems 2023, NeurIPS 2023, New Orleans, LA, USA, December 10 - 16, 2023}.

\bibitem[{Speechless(2023)}]{Speechless2023Huggingface}
Speechless. 2023.
\newblock \href {https://huggingface.co/uukuguy/speechless-codellama-34b-v2.0} {speechless-codellama-34b-v2.0}.

\bibitem[{Su et~al.(2024)Su, Ahmed, Lu, Pan, Bo, and Liu}]{RoFormer2024SuALPBL24}
Jianlin Su, Murtadha H.~M. Ahmed, Yu~Lu, Shengfeng Pan, Wen Bo, and Yunfeng Liu. 2024.
\newblock \href {https://doi.org/10.1016/J.NEUCOM.2023.127063} {Roformer: Enhanced transformer with rotary position embedding}.
\newblock \emph{Neurocomputing}, 568:127063.

\bibitem[{Touvron et~al.(2023{\natexlab{a}})Touvron, Lavril, Izacard, Martinet, Lachaux, Lacroix, Rozi{\`{e}}re, Goyal, Hambro, Azhar, Rodriguez, Joulin, Grave, and Lample}]{LLama2023ABS230213971}
Hugo Touvron, Thibaut Lavril, Gautier Izacard, Xavier Martinet, Marie{-}Anne Lachaux, Timoth{\'{e}}e Lacroix, Baptiste Rozi{\`{e}}re, Naman Goyal, Eric Hambro, Faisal Azhar, Aur{\'{e}}lien Rodriguez, Armand Joulin, Edouard Grave, and Guillaume Lample. 2023{\natexlab{a}}.
\newblock \href {https://doi.org/10.48550/ARXIV.2302.13971} {Llama: Open and efficient foundation language models}.
\newblock \emph{CoRR}, abs/2302.13971.

\bibitem[{Touvron et~al.(2023{\natexlab{b}})Touvron, Martin, Stone, Albert, Almahairi, Babaei, Bashlykov, Batra, Bhargava, Bhosale, Bikel, Blecher, Canton{-}Ferrer, Chen, Cucurull, Esiobu, Fernandes, Fu, Fu, Fuller, Gao, Goswami, Goyal, Hartshorn, Hosseini, Hou, Inan, Kardas, Kerkez, Khabsa, Kloumann, Korenev, Koura, Lachaux, Lavril, Lee, Liskovich, Lu, Mao, Martinet, Mihaylov, Mishra, Molybog, Nie, Poulton, Reizenstein, Rungta, Saladi, Schelten, Silva, Smith, Subramanian, Tan, Tang, Taylor, Williams, Kuan, Xu, Yan, Zarov, Zhang, Fan, Kambadur, Narang, Rodriguez, Stojnic, Edunov, and Scialom}]{LLama22023ABS230709288}
Hugo Touvron, Louis Martin, Kevin Stone, Peter Albert, Amjad Almahairi, Yasmine Babaei, Nikolay Bashlykov, Soumya Batra, Prajjwal Bhargava, Shruti Bhosale, Dan Bikel, Lukas Blecher, Cristian Canton{-}Ferrer, Moya Chen, Guillem Cucurull, David Esiobu, Jude Fernandes, Jeremy Fu, Wenyin Fu, Brian Fuller, Cynthia Gao, Vedanuj Goswami, Naman Goyal, Anthony Hartshorn, Saghar Hosseini, Rui Hou, Hakan Inan, Marcin Kardas, Viktor Kerkez, Madian Khabsa, Isabel Kloumann, Artem Korenev, Punit~Singh Koura, Marie{-}Anne Lachaux, Thibaut Lavril, Jenya Lee, Diana Liskovich, Yinghai Lu, Yuning Mao, Xavier Martinet, Todor Mihaylov, Pushkar Mishra, Igor Molybog, Yixin Nie, Andrew Poulton, Jeremy Reizenstein, Rashi Rungta, Kalyan Saladi, Alan Schelten, Ruan Silva, Eric~Michael Smith, Ranjan Subramanian, Xiaoqing~Ellen Tan, Binh Tang, Ross Taylor, Adina Williams, Jian~Xiang Kuan, Puxin Xu, Zheng Yan, Iliyan Zarov, Yuchen Zhang, Angela Fan, Melanie Kambadur, Sharan Narang, Aur{\'{e}}lien Rodriguez, Robert Stojnic, Sergey Edunov,
  and Thomas Scialom. 2023{\natexlab{b}}.
\newblock \href {https://doi.org/10.48550/ARXIV.2307.09288} {Llama 2: Open foundation and fine-tuned chat models}.
\newblock \emph{CoRR}, abs/2307.09288.

\bibitem[{Wang et~al.(2023)Wang, Ren, Zhou, Lu, Luo, Shi, Zhang, Song, Zhan, and Li}]{MathCoder2023ABS231003731}
Ke~Wang, Houxing Ren, Aojun Zhou, Zimu Lu, Sichun Luo, Weikang Shi, Renrui Zhang, Linqi Song, Mingjie Zhan, and Hongsheng Li. 2023.
\newblock \href {https://doi.org/10.48550/ARXIV.2310.03731} {Mathcoder: Seamless code integration in llms for enhanced mathematical reasoning}.
\newblock \emph{CoRR}, abs/2310.03731.

\bibitem[{Wang et~al.(2021)Wang, Wang, Joty, and Hoi}]{CodeT50034WJH21}
Yue Wang, Weishi Wang, Shafiq~R. Joty, and Steven C.~H. Hoi. 2021.
\newblock \href {https://doi.org/10.18653/V1/2021.EMNLP-MAIN.685} {Codet5: Identifier-aware unified pre-trained encoder-decoder models for code understanding and generation}.
\newblock In \emph{Proceedings of the 2021 Conference on Empirical Methods in Natural Language Processing, {EMNLP} 2021, Virtual Event / Punta Cana, Dominican Republic, 7-11 November, 2021}, pages 8696--8708. Association for Computational Linguistics.

\bibitem[{Wei et~al.(2022)Wei, Bosma, Zhao, Guu, Yu, Lester, Du, Dai, and Le}]{SelfInstruct2022WeiBZGYLDDL22}
Jason Wei, Maarten Bosma, Vincent~Y. Zhao, Kelvin Guu, Adams~Wei Yu, Brian Lester, Nan Du, Andrew~M. Dai, and Quoc~V. Le. 2022.
\newblock \href {https://openreview.net/forum?id=gEZrGCozdqR} {Finetuned language models are zero-shot learners}.
\newblock In \emph{The Tenth International Conference on Learning Representations, {ICLR} 2022, Virtual Event, April 25-29, 2022}. OpenReview.net.

\bibitem[{Wei et~al.(2023)Wei, Wang, Liu, Ding, and Zhang}]{Magicoder2023ABS231202120}
Yuxiang Wei, Zhe Wang, Jiawei Liu, Yifeng Ding, and Lingming Zhang. 2023.
\newblock \href {https://doi.org/10.48550/ARXIV.2312.02120} {Magicoder: Source code is all you need}.
\newblock \emph{CoRR}, abs/2312.02120.

\bibitem[{Wu et~al.(2024)Wu, Huang, Shi, Wang, Gao, Liu, Nan, Yuan, Zhang, Zhang, Du, Guo, Pu, Yin, Hu, and Chen}]{InverseCoder2024Yutong}
Yutong Wu, Di~Huang, Wenxuan Shi, Wei Wang, Lingzhe Gao, Shihao Liu, Ziyuan Nan, Kaizhao Yuan, Rui Zhang, Xishan Zhang, Zidong Du, Qi~Guo, Yewen Pu, Dawei Yin, Xing Hu, and Yunji Chen. 2024.
\newblock \href {https://doi.org/10.48550/ARXIV.2407.05700} {Inversecoder: Unleashing the power of instruction-tuned code llms with inverse-instruct}.
\newblock \emph{CoRR}, abs/2407.05700.

\bibitem[{Yang et~al.(2023)Yang, Xiao, Wang, Zhang, Bian, Yin, Lv, Pan, Wang, Yan, Yang, Deng, Wang, Liu, Ai, Dong, Zhao, Xu, Sun, Zhang, Liu, Ji, Xie, Dai, Fang, Su, Song, Liu, Ru, Ma, Wang, Liu, Lin, Nie, Guo, Sun, Zhang, Li, Li, Cheng, Chen, Zeng, Wang, Chen, Men, Yu, Pan, Shen, Wang, Li, Jiang, Gao, Zhang, Zhou, and Wu}]{BaiChuan2023ABS230910305}
Aiyuan Yang, Bin Xiao, Bingning Wang, Borong Zhang, Ce~Bian, Chao Yin, Chenxu Lv, Da~Pan, Dian Wang, Dong Yan, Fan Yang, Fei Deng, Feng Wang, Feng Liu, Guangwei Ai, Guosheng Dong, Haizhou Zhao, Hang Xu, Haoze Sun, Hongda Zhang, Hui Liu, Jiaming Ji, Jian Xie, Juntao Dai, Kun Fang, Lei Su, Liang Song, Lifeng Liu, Liyun Ru, Luyao Ma, Mang Wang, Mickel Liu, MingAn Lin, Nuolan Nie, Peidong Guo, Ruiyang Sun, Tao Zhang, Tianpeng Li, Tianyu Li, Wei Cheng, Weipeng Chen, Xiangrong Zeng, Xiaochuan Wang, Xiaoxi Chen, Xin Men, Xin Yu, Xuehai Pan, Yanjun Shen, Yiding Wang, Yiyu Li, Youxin Jiang, Yuchen Gao, Yupeng Zhang, Zenan Zhou, and Zhiying Wu. 2023.
\newblock \href {https://doi.org/10.48550/ARXIV.2309.10305} {Baichuan 2: Open large-scale language models}.
\newblock \emph{CoRR}, abs/2309.10305.

\bibitem[{Yu et~al.(2023)Yu, Zhang, Shang, Huang, Xu, Zhao, Hu, and Yin}]{WaveCoderABS231214187}
Zhaojian Yu, Xin Zhang, Ning Shang, Yangyu Huang, Can Xu, Yishujie Zhao, Wenxiang Hu, and Qiufeng Yin. 2023.
\newblock \href {https://doi.org/10.48550/ARXIV.2312.14187} {Wavecoder: Widespread and versatile enhanced instruction tuning with refined data generation}.
\newblock \emph{CoRR}, abs/2312.14187.

\bibitem[{Zhang et~al.(2023)Zhang, Chen, Liu, Liao, Gong, Yu, Li, and Wang}]{Survey2023Ziyin}
Ziyin Zhang, Chaoyu Chen, Bingchang Liu, Cong Liao, Zi~Gong, Hang Yu, Jianguo Li, and Rui Wang. 2023.
\newblock \href {https://doi.org/10.48550/ARXIV.2311.07989} {A survey on language models for code}.
\newblock \emph{CoRR}, abs/2311.07989.

\bibitem[{Zheng et~al.(2023)Zheng, Xia, Zou, Dong, Wang, Xue, Shen, Wang, Wang, Li, Su, Yang, and Tang}]{CodeGeeX2023ZhengXZDWXSW0LS23}
Qinkai Zheng, Xiao Xia, Xu~Zou, Yuxiao Dong, Shan Wang, Yufei Xue, Lei Shen, Zihan Wang, Andi Wang, Yang Li, Teng Su, Zhilin Yang, and Jie Tang. 2023.
\newblock \href {https://doi.org/10.1145/3580305.3599790} {Codegeex: {A} pre-trained model for code generation with multilingual benchmarking on humaneval-x}.
\newblock In \emph{Proceedings of the 29th {ACM} {SIGKDD} Conference on Knowledge Discovery and Data Mining, {KDD} 2023, Long Beach, CA, USA, August 6-10, 2023}, pages 5673--5684. {ACM}.

\bibitem[{Zheng et~al.(2024)Zheng, Zhang, Shen, Liu, Lin, Fu, Chen, and Yue}]{OpenCodeInterpreter2024ABS240214658}
Tianyu Zheng, Ge~Zhang, Tianhao Shen, Xueling Liu, Bill~Yuchen Lin, Jie Fu, Wenhu Chen, and Xiang Yue. 2024.
\newblock \href {https://doi.org/10.48550/ARXIV.2402.14658} {Opencodeinterpreter: Integrating code generation with execution and refinement}.
\newblock \emph{CoRR}, abs/2402.14658.

\bibitem[{Zhou et~al.(2023)Zhou, Wang, Lu, Shi, Luo, Qin, Lu, Jia, Song, Zhan, and Li}]{CSV2023ABS230807921}
Aojun Zhou, Ke~Wang, Zimu Lu, Weikang Shi, Sichun Luo, Zipeng Qin, Shaoqing Lu, Anya Jia, Linqi Song, Mingjie Zhan, and Hongsheng Li. 2023.
\newblock \href {https://doi.org/10.48550/ARXIV.2308.07921} {Solving challenging math word problems using {GPT-4} code interpreter with code-based self-verification}.
\newblock \emph{CoRR}, abs/2308.07921.

\bibitem[{Zhuo et~al.(2024)Zhuo, Vu, Chim, Hu, Yu, Widyasari, Yusuf, Zhan, He, Paul, Brunner, Gong, Hoang, Zebaze, Hong, Li, Kaddour, Xu, Zhang, Yadav, Jain, Gu, Cheng, Liu, Liu, Wang, Lo, Hui, Muennighoff, Fried, Du, de~Vries, and von Werra}]{BigCodeBench2024Terry}
Terry~Yue Zhuo, Minh~Chien Vu, Jenny Chim, Han Hu, Wenhao Yu, Ratnadira Widyasari, Imam Nur~Bani Yusuf, Haolan Zhan, Junda He, Indraneil Paul, Simon Brunner, Chen Gong, Thong Hoang, Armel~Randy Zebaze, Xiaoheng Hong, Wen{-}Ding Li, Jean Kaddour, Ming Xu, Zhihan Zhang, Prateek Yadav, Naman Jain, Alex Gu, Zhoujun Cheng, Jiawei Liu, Qian Liu, Zijian Wang, David Lo, Binyuan Hui, Niklas Muennighoff, Daniel Fried, Xiaoning Du, Harm de~Vries, and Leandro von Werra. 2024.
\newblock \href {https://doi.org/10.48550/ARXIV.2406.15877} {Bigcodebench: Benchmarking code generation with diverse function calls and complex instructions}.
\newblock \emph{CoRR}, abs/2406.15877.

\end{thebibliography}
